\documentclass[a4paper,fleqn]{cas-sc}

\usepackage[utf8]{inputenc}
\usepackage[authoryear]{natbib}

\usepackage{amsmath,amssymb,amsfonts}
\usepackage{graphicx}
\usepackage{subcaption}
\usepackage{caption}
\usepackage{booktabs}
\usepackage{multirow}
\usepackage{array}
\usepackage{tabularx}
\usepackage{float}
\usepackage{placeins}
\usepackage{hyperref}
\usepackage[dvipsnames]{xcolor}
\usepackage{longtable}
\usepackage{pdflscape}
\usepackage[most]{tcolorbox}
\usepackage{xcolor}
\usepackage{tabularx}
\usepackage{array}
\usepackage{svg}
\usepackage[most]{tcolorbox}
\usepackage{xcolor}
\usepackage{lineno}

\definecolor{promptheader}{RGB}{65,65,65}
\definecolor{promptback}{RGB}{248,248,248}
\definecolor{promptframe}{RGB}{75,75,75}

\newtcolorbox{promptbox}[1]{
    enhanced,
    breakable,
    colback=promptback,
    colframe=promptframe,
    boxrule=0.7pt,
    arc=1.5mm,
    left=4mm,
    right=4mm,
    top=3mm,
    bottom=3mm,
    title={#1},
    colbacktitle=promptheader,
    coltitle=white,
    fonttitle=\bfseries,
    toptitle=1.5mm,
    bottomtitle=1.5mm,
    lefttitle=4mm,
    titlerule=0pt
}

\definecolor{questionblue}{RGB}{20,105,220}
\definecolor{questionlight}{RGB}{245,248,252}
\definecolor{questiongray}{RGB}{115,125,140}

\definecolor{promptheader}{RGB}{65,65,65}
\definecolor{promptback}{RGB}{248,248,248}
\definecolor{promptframe}{RGB}{75,75,75}
\definecolor{answerblue}{RGB}{45,105,170}
\definecolor{answerback}{RGB}{236,243,250}

\newtcolorbox{questionexample}{
    enhanced,
    breakable,
    colback=promptback,
    colframe=promptframe,
    boxrule=0.6pt,
    arc=1mm,
    left=3mm,
    right=3mm,
    top=2mm,
    bottom=2mm,
    title={Example Diagnostic Question},
    colbacktitle=promptheader,
    coltitle=white,
    fonttitle=\bfseries,
    toptitle=1mm,
    bottomtitle=1mm,
    lefttitle=3mm,
    titlerule=0pt
}

\newtcolorbox{goldanswer}{
    enhanced,
    colback=answerback,
    colframe=answerblue,
    boxrule=0.8pt,
    arc=1mm,
    left=2mm,
    right=2mm,
    top=1.2mm,
    bottom=1.2mm,
    before skip=1mm,
    after skip=1mm
}

\usepackage{siunitx}
\begin{document}

\let\WriteBookmarks\relax
\def\floatpagepagefraction{1}
\def\textpagefraction{.001}

\shorttitle{}
\shortauthors{}


\title[mode=title]{Chain of Physics: Supporting Fault Diagnosis with Engineering-Inspired Large Language Model Reasoning}


\author[1]{Wenqi Zhou}
\ead{zhou@zhaw.ch}

\author[1]{Leonardo Baldo}
\cormark[1]
\ead{*Corresponding Author leonardo.baldo@zhaw.ch}

\author[2,3]{Kai Goebel}
\ead{kai.goebel@fragumglobal.com}

\author[1,4]{Manuel Arias~Chao}
\ead{aria@zhaw.ch}

\address[1]{Institute for Data Science, Zurich University of Applied Science, Winterthur, 8401, Switzerland}

\address[2]{Fragum Global, Mountain View, CA 94040, USA}

\address[3]{Luleå University of Technology, Luleå, 971 87, Sweden}

\address[4]{Faculty of Aerospace Engineering, Delft University of Technology, HS 2926 Delft, The Netherlands}


\maketitle

\begin{abstract}
Fault diagnosis in complex systems requires linking observed changes in physical quantities to the failure mechanisms that produce them. Experienced engineers perform this task by combining sensor information with knowledge of system behavior and fault propagation. Large Language Models (LLMs) can support this process, but their diagnostic reasoning depends on the availability of asset-specific engineering knowledge and on how this knowledge is represented and used during inference.
This work introduces Chain of Physics (CoP), an LLM-based fault-diagnosis framework built around a Constrained Knowledge Graph (CKG) generated automatically from asset-specific reference documentation. CoP addresses failure--sensor association from textual engineering knowledge. The CKG stores diagnostic chains connecting failure modes, changes in physical quantities, and monitoring sensors. During inference, semantically similar graph paths are retrieved for each question and used as both information and in-context examples for the reasoning process. Candidate failure modes are extracted from the question and analysed independently through dedicated LLM calls, which generate plausible failure-to-observation chains. A final decision module compares these chains with the available answer choices.
CoP is evaluated on the FailureSensorIQ benchmark across four industrial asset classes and three LLMs. Compared with direct prompting and conventional Retrieval-Augmented Generation (RAG), CoP improves diagnostic accuracy across all evaluated models, with gains ranging from 7.8 to 19.6 percentage points and from 11.7 to 18.2 percentage points, respectively. The results show that structured engineering knowledge can improve LLM-based diagnostic reasoning when graph retrieval is combined with in-context examples of fault propagation.
\end{abstract}

\section{Introduction}
\label{sec:intro}

Fault diagnosis in complex systems requires identifying the failure mechanisms responsible for an observed deviation from nominal behavior. Experienced engineers and technicians perform this task by combining available measurements with their knowledge of system behavior and reasoning about the physical quantities through which failure effects become observable \cite{iso17359_2018}. These relationships link observed sensor behavior to specific failure mechanisms, enabling Fault Detection and Isolation (FDI) and supplying diagnostic information for downstream health assessment and maintenance decisions in Prognostics and Health Management (PHM) frameworks \cite{jardine2006review}.
In industrial applications, the knowledge required for this process is generally distributed across technical manuals, reference documents, troubleshooting procedures, maintenance records, and accumulated expert experience \cite{kang2026maintenance}. Its interpretation and integration therefore remain largely dependent on experienced personnel, who must retrieve and reconcile information scattered across heterogeneous technical documents and information systems \cite{wang2024idskg,RAZAQ2026114908}. The problem considered in this work focuses specifically on diagnosis from engineering knowledge, not on the direct classification of measured sensor signals. Given a textual description of a failure mode or sensor observation, the objective is to recover the relevant failure--sensor association and the physical relationships that connect them. The task can therefore be viewed as knowledge association from engineering text, requiring the relationships among failure mechanisms, their physical effects, and the corresponding sensing modalities to be established from textual information.

The increasing interest in Large Language Models (LLMs) introduces the possibility of assisting engineers in this knowledge-intensive task \cite{10680717}. Through natural-language interaction, LLMs can process technical information, interpret diagnostic questions, and formulate possible explanations without requiring a dedicated model for each asset \cite{RAZAQ2026114908}. Despite the strong push towards Artificial Intelligence (AI) adoption and the proliferation of AI initiatives across industrial sectors, the integration of LLMs into fault diagnosis remains challenging. Conventional task-specific models generally require representative labelled data and retraining when new assets or failure conditions are introduced, limiting their applicability in industrial settings. LLMs offer greater flexibility, but often lack the asset-specific and practical engineering knowledge required to interpret abnormal conditions correctly. This limitation can lead to hallucinated or physically inconsistent diagnostic explanations, particularly when the model must establish relationships between failure mechanisms, their effects on system behavior, and the available sensor measurements \cite{ma2026FDRKG}. Their effective deployment therefore requires incorporating established engineering knowledge and guiding the reasoning process according to the physical relationships governing the system.

One of the most widely adopted approaches to incorporate external engineering knowledge into LLM inference is Retrieval-Augmented Generation (RAG) \cite{lukens2025manualRAG}. By retrieving relevant passages from technical documentation, RAG can provide information beyond that acquired during pre-training without requiring task-specific model training. Nevertheless, the availability of relevant information does not guarantee that the model can reconstruct the physical sequence linking a fault to its measurable effects. However, in fault diagnosis, these relationships are central to the reasoning process: a failure affects one or more physical quantities, its effects may propagate through the system, and the resulting changes are detected by specific sensors. Representing this sequence explicitly may therefore provide a more suitable basis for diagnostic reasoning than supplying unstructured technical information alone.
This type of explicit failure-to-observation reasoning has a long tradition in engineering diagnosis, including qualitative physics and model-based diagnosis \cite{deKleerWilliams1987}, Signed Directed Graph (SDG) representations \cite{kramer1987sdg}, and Failure Mode and Symptoms Analysis (FMSA) \cite{iso13379_1_2012}. These approaches similarly rely on structured relationships between failure mechanisms, their physical manifestations, and the observations through which they can be detected.
In recent years, Knowledge Graphs (KGs) have emerged as a possible means of representing this type of diagnostic structure through explicit entities and relationships \cite{ibrahim2024kgsllms}. For fault diagnosis, a KG can represent the progression from a failure mode to the physical changes and the measurements through which those changes become observable. Even more, the close integration of LLMs and KGs is increasingly being investigated for industrial fault diagnosis \cite{MARTINEZGIL2022944}. In fact, LLMs provide natural language processing and reasoning capabilities, while KGs organize domain knowledge into explicit entities and relationships. Their integration allows diagnostic systems to retrieve and use structured engineering knowledge during inference, supporting more informed and interpretable diagnostic reasoning. 
Still, the practical integration of LLMs and KGs into an industrial diagnostic framework presents two main difficulties \cite{pan2024roadmap}. First, manually constructing asset-specific knowledge representations requires considerable engineering effort, particularly when multiple systems and failure mechanisms are considered. Second, a graph containing relevant entities and generic relationships does not necessarily preserve the physical sequence required for diagnostic reasoning.

\begin{figure}
    \centering
    \includegraphics[width=\textwidth]{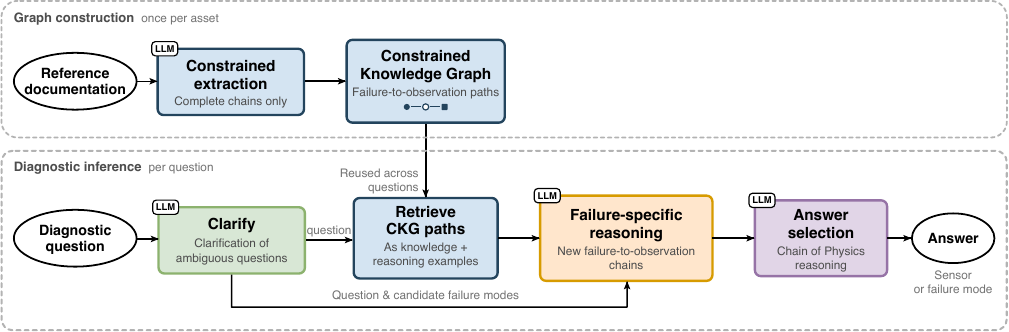}
    \caption{Overview of CoP. Once per asset, constrained extraction turns reference documentation into a CKG whose paths are complete failure-to-observation chains. For each diagnostic question, retrieved CKG paths provide both asset-specific knowledge and reasoning examples. These guide failure-specific reasoning, which produces new chains for the candidate failure modes. A separate answer-selection stage makes the decision based on those chains.}
    \label{fig:introduction_overview}
\end{figure}

Against this backdrop, this work introduces Chain of Physics (CoP) as an engineering-inspired framework for LLM-assisted fault diagnosis based on the CKG, aimed to tackle these exact two challenges. 
The methodology follows the way an experienced engineer approaches a diagnostic problem: plausible failure mechanisms are considered, their physical consequences are traced through the system, and the resulting effects are connected to the available measurements. This places CoP within the broader line of work that translates human reasoning procedures into explicit intermediate steps that can be reproduced algorithmically during LLM inference \cite{wei2022chain,gong2026sensorreasoning}. CoP organizes both knowledge extraction and diagnostic reasoning around the same failure-to-observation structure. Figure \ref{fig:introduction_overview} summarizes this logic, with the offline construction of reusable asset knowledge on the left and the question-specific reasoning process on the right. During graph construction, asset-specific reference documentation is processed automatically and candidate chains are retained only when the source material supports a failure mode, at least one change in a physical quantity, and a monitoring sensor. These elements are represented through a fixed set of relations describing the initial failure effect, its propagation through the system, and its observation by a sensor. The resulting CKG is therefore a diagnostic representation of the source knowledge in which each complete path corresponds to a physically structured failure-to-observation chain. Since the graph is generated once for each asset, the extracted knowledge can be reused across subsequent diagnostic questions.
During inference, relevant CKG paths are retrieved for the question under analysis and used in two ways. They provide asset-specific information about failure mechanisms and measurable effects, and they are supplied to the LLM as in-context examples of the reasoning structure to follow. Candidate failure modes extracted from the question are then analysed independently, allowing the model to generate new failure-to-observation chains that are not restricted to paths already contained in the graph. A separate decision stage gives the answer based on the generated chains. CoP therefore combines graph-grounded retrieval with in-context reasoning over a constrained engineering representation, while maintaining a clear separation between failure-specific reasoning and final answer selection.

The proposed approach is evaluated using FailureSensorIQ, an open-source benchmark addressing the relationships between industrial failure modes and monitoring sensors. The assessment considers two complementary tasks: identifying relevant sensors for a given failure mode and identifying possible failure modes from sensor evidence. Experiments are conducted using three LLMs on four asset classes: an aero gas turbine, an electric motor, a fan, and a power transformer. CoP is compared with direct LLM prompting and conventional RAG based on the same reference documentation. An ablation study further examines the contribution of question clarification and the separate diagnostic-reasoning and decision stages. The results show that CoP improves diagnostic accuracy across all evaluated models, supporting the use of explicitly structured engineering knowledge to assist fault diagnosis without task-specific model training.
In summary, the main contributions of this work are:

\begin{itemize}

    \item An automatic procedure for constructing an asset-specific CKG from technical reference documentation. The graph is restricted to complete diagnostic chains linking failure modes, changes in physical quantities, and monitoring sensors through a fixed set of relations.

    \item A diagnostic inference framework that uses retrieved CKG paths as both engineering knowledge and in-context reasoning exemplars. These paths guide the generation of failure-specific chains without requiring the target failure mode to be explicitly represented in the retrieved graph paths.

    \item A staged reasoning procedure that separates question clarification, failure-mode extraction, graph retrieval, failure-specific reasoning, and final answer selection, allowing the physical reasoning process to be performed independently from the multiple-choice decision.

    \item An evaluation on the FailureSensorIQ benchmark across four industrial asset classes and three LLMs, including comparisons with direct prompting and conventional RAG and an ablation analysis of the main inference components.

\end{itemize}

The remainder of the paper is organised as follows. Section~\ref{sec:related_work} reviews related work on LLM-based fault diagnosis, KGs, RAG, and prompt-based reasoning. Section~\ref{sec:methodology} presents the proposed Chain of Physics framework, including the offline construction of the CKG and the online diagnostic inference procedure. Section~\ref{sec:dataset} introduces the FailureSensorIQ benchmark and the experimental setting. Section~\ref{sec:results} presents and discusses the results. Finally, Section~\ref{sec:conclusions} summarises the main findings and outlines directions for future work.

\section{Related Works}
\label{sec:related_work}

The proposed framework lies at the intersection of three research directions: (i) the use of external knowledge during LLM inference, (ii) the representation of domain knowledge through graph-based structures, and (iii) the use of in-context examples to guide reasoning. These areas have developed largely in parallel, but their combination is particularly relevant for fault diagnosis, where both asset-specific knowledge and the structure of the reasoning process matter. The following paragraph briefly introduces representative developments from the broader Machine Learning (ML) literature that are relevant to the methodological building blocks considered in this work, including RAG, graph-based knowledge access, and In-Context Learning (ICL). After this broader methodological overview, the second part of the section reviews works more directly connected to PHM and fault diagnosis. The purpose of the next paragraph is then not to provide a comprehensive review of current ML research, but to establish the technical background and state of the art needed to position the proposed framework.  For a broader overview of the interaction between LLMs and structured knowledge, the interested reader is referred to Pan et al. \cite{pan2024roadmap}, who provide a comprehensive review of KG-enhanced LLMs, LLM-augmented KGs, and complete LLM--KG frameworks.

A central limitation of LLMs in knowledge-intensive applications is their dependence on information acquired during pre-training. This has motivated the development of methods that provide models with external knowledge at inference time. RAG, introduced by Lewis et al. \cite{lewis2020rag}, is one of the most established approaches in this direction. Conventional RAG retrieves relevant passages from an external document collection and supplies them to the model as additional context for question answering or text generation.
Beyond unstructured text retrieval, subsequent work has investigated how external knowledge can be represented and accessed in a more structured form. Jiang et al. \cite{jiang2023structgpt} proposed StructGPT, which enables LLMs to interact with structured data through dedicated interfaces. Graph-based approaches then extended this idea by exploiting explicit entities, relations, and paths during retrieval and reasoning. Wang et al. \cite{wang2024kgp}, for example, introduced Knowledge Graph Prompting, where graph construction and traversal are used to retrieve supporting evidence from multiple documents. Luo et al. \cite{luo2024rog} developed Reasoning on Graphs (RoG), in which relation paths are used to guide the retrieval of graph-grounded evidence. More recently, Ma et al. \cite{ma2025tog2} proposed Think-on-Graph 2.0, which combines structured graph retrieval with unstructured textual information. These approaches show how graph structure can be used to organize external knowledge and make relational information directly available during LLM inference.
A related line of research concerns ICL, where examples included in the input prompt guide the model in performing a task without modifying its parameters. The choice of these examples can substantially affect model behavior, motivating retrieval-based approaches in which demonstrations are selected dynamically for each input \cite{rubin2022retrievedicl,liu2022incontextexamples}. This principle has also been applied to graph-structured knowledge. Wei et al. \cite{wei2023kicgpt}, for instance, proposed KICGPT, where structural knowledge retrieved from a KG is encoded into demonstrations that guide the LLM during knowledge-graph completion. In this setting, retrieval determines both the information supplied to the model and the examples used to shape its reasoning process.


\begin{table}[htbp]
\centering
\caption{Representative studies grouped by methodological similarity and their relationship to CoP.}
\label{tab:litreview-condensed}

\renewcommand{\arraystretch}{1.15}
\setlength{\tabcolsep}{4pt}
\footnotesize

\begin{tabularx}{\textwidth}{
    >{\raggedright\arraybackslash}p{2.45cm}
    >{\raggedright\arraybackslash}p{3.35cm}
    >{\raggedright\arraybackslash}X
    >{\raggedright\arraybackslash}X
}
\toprule
\textbf{Group} &
\textbf{Representative studies} &
\textbf{Key characteristic} &
\textbf{Distinction from CoP} \\
\midrule

Reasoning procedure without persistent KG
&
Gong et al. \cite{gong2026sensorreasoning}
&
Structured prompting and agent-based architectures, including multi-agent debate and planning.
&
No reusable, offline-built engineering-knowledge representation. \\

\addlinespace

KG-based symptom / solution retrieval
&
Lin and Miao \cite{electronics14214180}; Nie et al. \cite{NIE2026111879}; Liu et al. \cite{liu2025CNC}
&
Graph retrieval links symptoms and faults to causes and maintenance actions.
&
Do not explicitly impose CoP's constraining schema. \\

\addlinespace

Graph-grounded retrieval and reasoning
&
Gu et al. \cite{gu2026causalgraph}; Xie et al. \cite{s24248086}; Miao et al. \cite{MIAO2026114855}
&
Retrieved causal paths or subgraphs provide structured, query-specific context to support subsequent LLM reasoning.
&
Evaluated in individual application domains, without a common cross-asset benchmark or not reproducible. \\

\addlinespace

Causal / mechanism-explicit KG
&
Guo et al. \cite{guo2026fault}
&
Weighted causal graph of fault-propagation paths derived from monitoring data.
&
Constructed from operational measurements instead of reference technical documentation. \\

\addlinespace

Automatic knowledge construction from documentation
&
Zhou et al. \cite{ZHOU2026100137}; Wang et al. \cite{wang2026graphragkm}; Marandi et al. \cite{marandi2025complex}
&
Automatic extraction of structured knowledge from technical documents. Marandi et al. also reuse the resulting representation for graph traversal and LLM-based reasoning.
&
Use different ontologies or functional representations, without CoP's predefined constraining schema. \\

\addlinespace

Comparable evaluation design
&
Ma et al. \cite{ma2026FDRKG} (FDRKG-LLM)
&
Comparison of a KG-grounded LLM with direct prompting and multiple conventional RAG baselines.
&
Single equipment class, with open-ended generation metrics instead of multi-asset multiple-choice accuracy. \\

\bottomrule
\end{tabularx}
\end{table}

In this sense, the use of LLMs in PHM has received increasing attention, with recent studies investigating their use across different stages of the PHM process. In their paper, Li et al. \cite{li2024foundationPHM} reviewed the development of large-scale foundation models for PHM, outlining their potential applications, methodological requirements, and remaining challenges. Chen et al. \cite{chen2025largemodels} reviewed large models for machine monitoring and fault diagnostics, while Tsallis et al. \cite{tsallis2026survey} provided a systematic review of large language and foundation models for machinery health monitoring. Razaq et al. \cite{RAZAQ2026114908} conducted a systematic literature review of 37 studies combining LLMs and KGs for fault diagnosis, identifying two dominant integration paradigms (i.e., automated KG construction and KG-grounded (GraphRAG-style) reasoning) and reporting that current work concentrates almost exclusively on rotating machinery, steel manufacturing, and Computer Numerical Control (CNC) equipment, highlighting limited multilingual coverage and persistent challenges concerning the explainability of diagnostic reasoning. Other contributions have investigated broader frameworks intended to support multiple PHM tasks. Cha et al. \cite{cha2025agent}, for instance, proposed an LLM-based autonomous agent that orchestrates data preprocessing, model selection, training, and optimization for fault classification and Remaining Useful Life (RUL) prediction. Tao et al. \cite{tao2025phmlm} introduced the concept of a PHM Large Model (PHM-LM), discussing its underlying paradigms and the technical challenges associated with its development and application.

Alongside these broader frameworks, a growing body of research addresses individual PHM functions, including fault detection and isolation, prognostics, and maintenance decision support. Fault diagnosis is particularly relevant for knowledge-based LLM applications, as the interpretation of abnormal conditions often requires combining sensor evidence with information contained in technical manuals, troubleshooting procedures, and maintenance records. Such information is frequently expressed in natural language, making it accessible to LLM-based reasoning and question-answering approaches.

Before considering recent LLM-based approaches to this problem, it is worth recalling that explicit representations of physical behaviour have a long history in engineering fault diagnosis. Classical work on qualitative physics represented system behaviour through qualitative changes in physical quantities and causal relationships, providing a basis for reasoning about how physical effects propagate through engineered systems \cite{deKleerWilliams1987}. Closely related model-based diagnostic approaches infer possible faults by comparing manifested system behaviour with the behaviour predicted from explicit models of component structure and function. In process diagnosis, causal and SDG representations have similarly been used to describe qualitative relationships among process variables and support the propagation of fault effects \cite{kramer1987sdg}. FMSA provides a further connection from the condition-monitoring domain, relating identified failure modes to their symptoms and to the monitoring techniques suitable for their detection \cite{iso13379_1_2012}.

The use of LLMs for fault diagnosis and diagnostic reasoning is therefore the research direction most closely related to the present work and requires a deeper look. The main axis along which different approaches can be distinguished is the way engineering knowledge is made available to the model and subsequently used: (i) some methods rely primarily on knowledge acquired during pre-training, (ii) others retrieve relevant passages from technical documentation, (iii) while a third group organizes domain information into structured representations such as KGs. 

Approaches based mainly on pretrained model knowledge use prompting, reasoning strategies, or agent-based decomposition without introducing an external persistent representation of asset-specific engineering knowledge \cite{gong2026sensorreasoning}. As already mentioned, their main limitation is that diagnostic performance depends on information already encoded in the model parameters, which may be incomplete, too generic, or inconsistent with the the physics of the specific asset under analysis. This becomes particularly critical when diagnosis requires detailed relationships between failure mechanisms, local changes in physical quantities, and the sensors through which these effects become observable.
A second group augments the LLM with unstructured technical information retrieved from manuals, maintenance documents, or troubleshooting material. Retrieval can provide relevant asset-specific evidence and reduce the dependence on pre-training knowledge. However, the retrieved passages remain expressed in natural language and the model must reconstruct the diagnostic relationships contained in them during inference. Relevant evidence may therefore be available in the context without explicitly preserving the sequence through which a failure produces physical effects and eventually becomes observable at a sensor. Lukens et al. \cite{lukens2025manualRAG}, for example, investigated the use of technical manuals within a RAG framework for fault diagnosis, providing a direct example of document-based knowledge augmentation in PHM.
Finally, structured approaches address this limitation by representing engineering knowledge explicitly through entities and relations. In these methods, knowledge extraction and representation become part of the diagnostic framework, allowing information about components, failure mechanisms, symptoms, and monitoring variables to be reused during inference. However, the usefulness of the resulting graph depends strongly on the adopted representation. Generic entity--relation graphs can store relevant domain information without necessarily preserving the physical progression required to reason from a failure mechanism to its measurable effects. A further refinement of this third category is provided by constrained knowledge representations, in which the admissible relations between entities are defined a priori according to the semantics of the target problem. In this case, the graph structure itself restricts how knowledge can be represented and connected, reducing arbitrary relation types and enforcing paths that follow a predefined engineering logic. For diagnostic applications, such constraints can be used to preserve the progression from a failure mechanism, through changes in physical quantities, to the sensor that observes their effects.

The next paragraph focuses on applications of this last paradigm, which is particularly relevant and promising for fault-diagnosis reasoning.
Among the applications combining LLMs and structured diagnostic knowledge, Liu et al. \cite{liu2024aviationKG} proposed a joint KG--LLM model for fault localization in aviation assembly. The method incorporates KG embeddings through prefix-tuning and uses knowledge retrieval to support the identification of fault causes and troubleshooting solutions. Ma et al. \cite{ma2026FDRKG} introduced FDRKG-LLM, combining a fault diagnostic reasoning KG with an LLM to support equipment diagnosis and maintenance decisions. Their evaluation compares the proposed framework with direct prompting and conventional RAG configurations on a self-constructed diagnostic database, reporting improvements in text-generation metrics. Liu et al. \cite{liu2025CNC} and Nie et al. \cite{NIE2026111879} investigated related approaches for CNC equipment, combining a domain KG, retrieval, and engineering information to support interactive fault diagnosis. Guo et al. \cite{guo2026fault} combined constrained causal discovery with LLM-enhanced knowledge-graph completion for rotating machinery, deriving a weighted causal directed acyclic graph that traces fault-propagation paths from monitoring data. A comparably large empirical gain was reported by Lin and Miao \cite{electronics14214180}, who improved automotive fault-diagnosis accuracy using a dual-mode knowledge-graph and LLM retrieval scheme. These studies establish the use of structured engineering knowledge in LLM-assisted industrial diagnosis, with different approaches to knowledge integration and model adaptation.

Since the integration of LLMs with structured engineering knowledge appears especially promising for scalable and reproducible fault diagnosis, the next question is: how are diagnostic knowledge representations created? For instance, Liao et al. \cite{liao2025roboticKG} investigated LLM-assisted extraction of fine-grained information from robotic maintenance texts for KG construction, and comparable automatic pipelines have been proposed for automotive \cite{ZHOU2026100137} and aviation \cite{LI2026101039} components. Wang et al. \cite{wang2026graphragkm} proposed GraphRAG-KM, an automated framework that converts industrial technical documentation into ontology and conceptual models through document parsing, graph-based retrieval, LLM extraction, and clustering. Marandi et al. \cite{marandi2025complex} proposed a framework for nuclear power plant diagnostics in which a hierarchical functional model is automatically constructed from technical documentation and encoded into a KG. The resulting representation supports both structured graph traversal and LLM-based interpretive reasoning through Graph-RAG. These contributions address the preparation and reuse of engineering knowledge, including its integration into diagnostic inference.

Other approaches investigate how retrieved knowledge can be used to structure the reasoning process itself. Gong et al. \cite{gong2026sensorreasoning} proposed a multi-agent framework that generates diagnostic reasoning blueprints and adapts the subsequent decision process according to question complexity. Evaluated on FailureSensorIQ \cite{constantinides2025FailureSensorIQ}, the framework improves Llama-3.1-8B accuracy from 36.5\% to 54.6\%. Gu et al. \cite{gu2026causalgraph} combined domain knowledge and data-driven information to construct a unified causal graph for fault diagnosis in distributed computing systems, supplying retrieved paths as in-context reasoning exemplars. Xie et al. \cite{s24248086} and Miao et al. \cite{MIAO2026114855} investigated related graph-grounded reasoning approaches for sensor networks and train bogies, respectively. Gao et al. \cite{gao2026causalFD} proposed LLM-FDQA, a KG-enhanced framework for industrial fault diagnosis that combines semantic extraction, two-stage entity linking, and multi-hop causal reasoning. The method retrieves and validates symptom--cause--solution chains to support diagnostic question answering. Evaluated on a proprietary CNC fault-diagnosis dataset, it reports an end-to-end diagnostic accuracy of 94.04\%. These studies show how graph-derived evidence can be used both as retrieved knowledge and as structure for subsequent diagnostic reasoning.

Reproducibility remains a practical limitation when these approaches are compared. Several studies rely on proprietary datasets or provide insufficient implementation detail for a controlled reimplementation of the complete reasoning pipeline. The paper from Gong et al. \cite{gong2026sensorreasoning} is among the few works evaluated on the public FailureSensorIQ benchmark whose reasoning architecture is described in sufficient detail to reproduce. This makes their framework particularly suitable for the additional controlled comparison reported in Section~\ref{subsec:no_ckg}.


The present work combines these research directions within a single diagnostic framework. From the perspective of classical diagnostic methods discussed above, the CKG follows the same general principle of representing explicitly the relationships between a failure and its observable manifestations, while adopting a structure tailored to LLM-based reasoning. Its paths contain failure modes, directional changes in physical quantities, and monitoring sensors connected through a constrained sequence of relations. Moreover, the methodology is shaped by the way an engineer approaches a diagnostic problem, by following the physical relationships that connect a failure mechanism to the changes it produces in the system and to the measurements through which those changes become observable. CoP automatically constructs an asset-specific CKG from reference documentation, retaining chains that connect failure modes, changes in physical quantities, and monitoring sensors. During inference, query-conditioned graph retrieval recovers complete CKG paths that provide both asset-specific engineering knowledge and in-context exemplars of failure-to-observation reasoning. Candidate failure modes are extracted independently from the diagnostic question and analysed through separate LLM calls using the retrieved chains as reasoning examples. The resulting failure-specific chains are then passed to a separate Judge module that selects the final answer. CoP therefore combines graph-grounded retrieval with retrieval-conditioned in-context reasoning and a distinct final decision stage.

\section{Methodology}
\label{sec:methodology}

Evaluating diagnostic reasoning is not straightforward. In realistic applications, the available evidence, operating conditions, and level of diagnostic detail can vary substantially, making direct comparison between reasoning approaches difficult. A common way of reducing this variability is to formulate the task as question answering, where the information required to reach the answer is controlled and the final response can be evaluated against a known target. In Knowledge Graph Question Answering (KGQA), this setting is commonly distinguished between simple questions, whose answer can be obtained from a single relation, and more complex multi-hop questions, which require combining several entities and relations before reaching the target answer \cite{song2023complexkgqa,peng2023knowledgegraphs}. This formulation provides a practical way of evaluating whether a model can recover isolated associations or compose a sequence of intermediate relations and it is one of the standard methodologies to test and compare reasoning algorithm in ML also thanks to the various published ML datasets.

A similar setting is commonly used for diagnostic reasoning. A failure-mode-to-sensor question may be answered through a direct association when the relevant relation is explicit, or it may require reconstructing a sequence in which a failure affects one or more physical quantities before its effect becomes observable at a sensor. FailureSensorIQ \cite{constantinides2025FailureSensorIQ} is, to the best of the authors' knowledge, one of the few and the most comprehensive publicly available benchmarks designed specifically to evaluate this type of reasoning across industrial assets. The dataset formulates diagnosis as multiple-choice question answering and tests both directions of the failure-mode--sensor relationship: identifying the sensors relevant to a given failure mode and inferring plausible failure modes from abnormal sensor observations.
Accordingly, the task considered here differs from conventional fault classification based on measured sensor data. FailureSensorIQ evaluates diagnostic association from textual information. Given a failure mode, the model must identify the sensing modality most relevant to its detection, or, in the inverse direction, determine which failure modes are consistent with a stated sensor observation. An example of a Failure Mode to Sensor (FM2Sensor) question is reported below, where the model receives a diagnostic stem and a fixed set of candidate sensor classes and must identify the right (gold) response.

\begin{questionexample}

\textbf{Stem.}
In an aero gas turbine, when air inlet blockage occurs, which sensor from the choices is most critical in detecting the occurrence of the failure event?

\medskip

\begin{tabularx}{\linewidth}{
    >{\bfseries}p{7mm}
    X
    >{\raggedleft\arraybackslash}p{15mm}
}
A & Gas generator temperature & \\[0.8mm]
B & Power turbine temperature & \\[0.8mm]
C & Oil debris & \\[0.8mm]

\rowcolor{answerback}
\textcolor{answerblue}{D} &
\textbf{Compressor temperature} &
\textbf{\textcolor{answerblue}{Gold}} \\[0.8mm]

E & Pressure / pressure ratio &
\end{tabularx}

\end{questionexample}

The example also illustrates a practical characteristic of the benchmark: diagnostic questions are not always expressed with fully standardised or technically precise wording. Since FailureSensorIQ is generated through a template-based procedure, this should not be interpreted as reproducing real industrial conversations. However, redundant formulations, broad sensor names, and ambiguous expressions are present in the benchmark and resemble the linguistic imprecision that can occur in technical diagnostic requests. This makes the dataset suitable for evaluating a pipeline that must first establish a clear interpretation of the question before reasoning about the underlying failure.

\begin{figure}
    \centering
    \includegraphics[width=\textwidth]{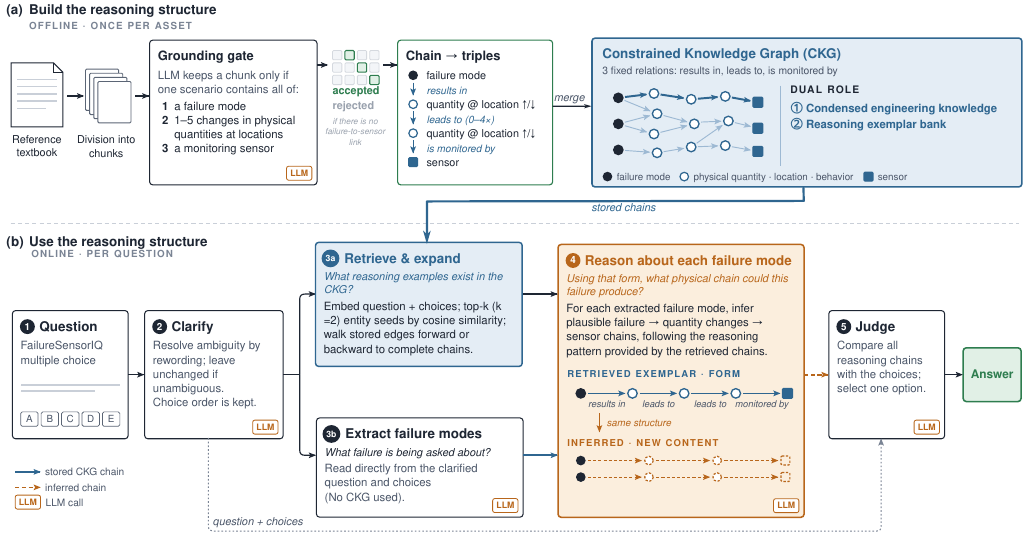}
    \caption{Complete CoP methodology. The framework consists of an offline stage, performed once for each asset, and an online stage executed for each diagnostic question. Offline, the asset-specific reference documentation is divided into chunks and processed through a grounding gate. The accepted chains are converted into triples using the fixed relations \textit{results in}, \textit{leads to}, and \textit{is monitored by}, and subsequently merged into the CKG. Online, the clarified question is used independently to retrieve and expand relevant CKG paths and to extract the failure modes under consideration. The retrieved paths guide the generation of new diagnostic chains for each failure mode, after which the Judge maps the resulting reasoning to the available answer choices.}
    \label{fig:mainFlowchart}
\end{figure}

CoP organizes diagnostic reasoning around the physical consequences expected from each candidate failure mechanism. The procedure follows the way an engineer approaches diagnosis by tracing the changes that a possible cause would produce in the system and identifying the measurements through which those changes become observable. For failure-mode-to-sensor questions, the reasoning proceeds from a specified failure to its observable effects. For sensor-to-failure-mode questions, candidate causes are assessed through their expected effects and their relevance to the stated observations. The method comprises an offline stage, performed once for each asset, in which reference documentation is converted into the CKG, and an online stage, executed for each diagnostic question, in which retrieved chains support failure-specific reasoning and subsequent answer selection. Figure~\ref{fig:mainFlowchart} visualizes the modularity: the upper part constructs the reusable asset-specific CKG, while the lower part shows how that representation is used for each diagnostic question.

All LLM calls in CoP use dedicated instruction templates designed following established prompt-design principles: the instructions define the task of each module, the context made available to it, the permitted operations, and the required output structure as indicated in \cite{schulhoff2024prompt,bsharat2023principled}. Complex diagnostic reasoning is decomposed into separate calls, retrieved failure-to-observation chains are supplied as in-context demonstrations where appropriate \cite{wei2022chain}, and structured outputs are used to constrain the information exchanged between stages. The complete prompts for all calls are reported in Appendix~\ref{app:prompts}.

\subsection{Offline step: CKG generation from Reference Knowledge}
\label{subsec:offline_CKG}

The offline stage is performed once for each asset and converts the corresponding reference documentation into the CKG used during online inference. As illustrated in Figure~\ref{fig:mainFlowchart}, the procedure comprises different operations: document parsing and segmentation, extraction of grounded failure-to-observation chains, and aggregation of the accepted chains into the asset-specific CKG.

\subsubsection{Document parsing and segmentation}
\label{subsubsec:document_parsing}

The reference documentation is first converted from PDF to Markdown using Docling \cite{docling2024}. The parsing pipeline preserves the document structure and enables the extraction of tables and mathematical expressions, while Optical Character Recognition (OCR) can be enabled when required by the source document. The resulting Markdown representation is then divided into text units of approximately 2,000 characters, which are processed independently in the following stage. This size has been selected following the commonly recommended starting point for fixed-size chunking \cite{microsoft_chunking} and lies within the range shown to benefit retrieval on technical, context-dependent documents \cite{bhat2025chunk}. 

This preprocessing step provides a common textual representation of the source material while keeping the subsequent LLM calls local to relatively short portions of the document. Each unit is treated independently, so the generation of a diagnostic chain is based only on the information contained in that portion of the reference text.

\subsubsection{Grounded diagnostic-chain extraction}
\label{subsubsec:chain_extraction}

Each text unit is passed to a dedicated LLM call using a structured output schema designed specifically for the diagnostic representation adopted in CoP. The extraction prompt is reported in Appendix~\ref{app:prompt_offline}. For each unit, the model is asked to extract one coherent monitoring scenario composed of a failure mode, an ordered sequence of changes in physical quantities, and the sensor that observes the final quantity. Each intermediate step contains the physical variable, its location on the asset when available, and the direction of change, restricted to \textit{increases} or \textit{decreases}. Between one and five physical-variable changes are allowed in each chain to avoid excessively long chains.

The extraction prompt acts as a grounding gate. A chain is retained only when the text unit supports the complete relationship between a failure mode, at least one physical-variable change, and a monitoring sensor within the same scenario. General theoretical descriptions, historical information, design equations, or passages without an explicit failure-to-monitoring relationship are therefore discarded. The model is also instructed to use only information contained in the current text unit, without introducing additional physical propagation steps. When several diagnostic scenarios are present, the dominant one is selected.

The structured LLM output is subsequently checked before graph generation. A candidate chain is rejected if the failure mode or sensor is missing, if no physical-variable step is available, or if any step does not contain a physical variable. Repeated node descriptions within the same chain are also rejected, since their direct conversion into the graph representation could introduce an artificial cycle. This second validation therefore complements the constraints imposed through the extraction prompt.
For an accepted result, the resulting sequence has the form
$$
\text{failure mode}
\rightarrow
\text{physical change}_1
\rightarrow
\cdots
\rightarrow
\text{physical change}_n
\rightarrow
\text{sensor}, \text{where} \ n \in [1, 5].
$$
Throughout the chain, each physical-change node combines the variable, its location when reported in the source, and its direction of change.

\subsubsection{Graph construction and aggregation}
\label{subsubsec:graph_construction}

Each accepted chain is converted directly into a small graph using the three relations defined by the CKG schema. The transition from the failure mode to the first physical effect is assigned the relation \textit{results in}; intermediate propagation steps are connected through \textit{leads to}; and the final physical effect is connected to the corresponding sensor through \textit{is monitored by}. The graph structure is therefore generated directly from the ordered chain returned by the extraction stage.
A separate graph is produced for every accepted text unit. Once all units associated with an asset have been processed, the resulting graphs are aggregated by taking the union of their entities, edge types, and relations using KGGen \cite{mo2025kggen}. The outcome is a single asset-specific CKG containing the failure-to-observation chains supported by the reference documentation.

This construction procedure imposes the same reasoning structure across all retained knowledge. Each complete path starts from a failure mechanism, follows the physical quantities through which its effects propagate, and terminates at the sensor through which the final effect becomes observable. The intuitive structure of the CKG allows experts to cross check and edit the information easily. The CKG can consequently be reused during online inference both as a condensed representation of asset-specific engineering knowledge and as a repository of complete diagnostic reasoning examples.

The extracted chains used in the evaluation were reviewed by a Subject-Matter Expert (SME) and corrected where required before online inference. The constrained representation makes this review direct: the initiating condition, every physical-variable change, its direction, and the monitoring sensor are exposed as an ordered path whose individual steps can be checked against engineering knowledge. The resulting knowledge bases remain compact enough for complete expert inspection. Section~\ref{subsec:ckg_results} and Figure~\ref{fig:ckg_turbofan_excerpt} show an excerpt of the resulting CKG and illustrate how the stored chains can be followed and verified relation by relation.

\subsection{Online step: Inference on a Specific Question}
\label{subsec:online_inference}


The online stage, reported in the lower half of Figure~\ref{fig:mainFlowchart}, is executed independently for each diagnostic question using the CKG previously constructed for the corresponding asset. Its purpose is to combine the information contained in the question with the reasoning patterns stored in the graph, while keeping knowledge retrieval, failure-mode identification, physical reasoning, and answer selection as separate operations.

\subsubsection{Question clarification}
\label{subsubsec:clarification}

The first operation is a clarification step performed by a dedicated LLM call using the prompt reported in Appendix~\ref{app:prompt_clarification}. The rationale follows a common pattern in human diagnostic reasoning: when a problem statement is unclear or ambiguous, an engineer first reformulates it to obtain a more precise interpretation before analysing the underlying failure mechanisms. The model receives the original question together with all candidate answers and determines whether the wording is sufficiently clear for the subsequent reasoning stages. If no ambiguity is identified, the original question and options are retained unchanged. Otherwise, the model reformulates the question and, where required, the answer options while preserving their number, order, meaning, asset, and diagnostic intent. The clarification module is explicitly instructed not to select an answer or introduce failure modes or sensors that are not implied by the original item. If the structured output cannot be parsed, or if the number of returned options differs from the original one, the pipeline falls back to the unmodified question.

The resulting question and options are used by all subsequent modules. The clarification stage therefore acts only on the interpretation of the diagnostic problem and does not access the CKG. Figure~\ref{fig:cop_reasoning_example} later shows a question that is left unchanged, while Appendix~\ref{app:clarify_example} reports a complementary worked example in which ambiguity is detected and the question is reformulated before inference.

\subsubsection{CKG retrieval and failure-mode extraction}
\label{subsubsec:retrieval_extraction}

After clarification, the pipeline separates into two parallel operations. 

The first retrieves relevant reasoning examples from the asset-specific CKG. A retrieval query is formed by concatenating the clarified question and all answer options. This query is embedded and compared through cosine similarity with cached embeddings of the failure-mode and sensor entities contained in the CKG. The top-$k$ highest-scoring entities are retained as retrieval seeds, where $k=2$ in the experiments reported in this work.
The selected seeds are then expanded along the graph to recover complete failure-to-sensor chains. This operation corresponds to graph traversal, i.e., following the stored relations from a selected node to reconstruct a complete path. When the retrieved seed is a failure mode, the algorithm follows the relations forward until a monitoring sensor is reached. When the seed is a sensor, the relations are followed backward until the corresponding failure mode is reached. The path expansion follows the constrained graph structure and recovers complete chains of the general form
$$
\text{failure mode}
\rightarrow
\text{physical effect}
\rightarrow \cdots \rightarrow
\text{sensor},
$$
using the relations \textit{results in}, \textit{leads to}, and \textit{is monitored by}. Duplicate paths are removed before the retrieved chains are supplied to the reasoning stage.

In parallel, a second LLM call extracts the failure modes mentioned or implied by the clarified question and its answer options using the prompt reported in Appendix~\ref{app:prompt_failure_extraction}. This operation is independent of the CKG. The extracted failure modes are not required to correspond to nodes already present in the graph. The result is a list of candidate failure modes defining what must subsequently be analysed.

This separation assigns different roles to the two branches to define what to reason for and how to reason. The question determines the failure modes under consideration, while the CKG provides asset-specific examples of how failure effects propagate toward measurable quantities. The target failure mode therefore does not need to be explicitly represented in the retrieved graph paths.

\subsubsection{Retrieval-conditioned in-context reasoning}
\label{subsubsec:retrieval_conditioned_reasoning}

The reasoning stage follows an ICL formulation. As already mentioned, in ICL, examples placed in the input context guide the model in performing the target task without updating its parameters \cite{rubin2022retrievedicl}. Retrieval-based variants select these demonstrations dynamically according to the current input, allowing the context presented to the model to change from one query to another \cite{liu2022incontextexamples}. Related graph-based approaches have used structured KG knowledge as in-context demonstrations to guide LLM inference \cite{wei2023kicgpt}.

CoP adopts retrieval-conditioned few-shot prompting at the level of diagnostic reasoning. Complete CKG chains are dynamically retrieved for each question and supplied as demonstrations of the failure-to-observation structure to follow. The complete CKG chains retrieved in the previous step are inserted into the prompt as examples of the expected failure-to-observation reasoning structure. They therefore provide both asset-specific engineering context and demonstrations of the sequence
"$
\text{failure mode}
\rightarrow
\text{physical quantity changes}
\rightarrow
\text{sensor}
$".

A separate LLM call is performed for every extracted failure mode using the prompt reported in Appendix~\ref{app:prompt_reasoning}. This separation mirrors the engineering reasoning process adopted throughout CoP. Each candidate failure is examined through its expected physical consequences before the final diagnostic comparison. The model receives the asset name, the target failure mode, the retrieved CKG paths, and the chain format associated with the graph. Its task is to trace plausible changes in physical quantities and identify the sensors capable of observing them. The reasoning module does not receive the multiple-choice question or its answer options directly; the assessment of the generated chains against the diagnostic question is assigned to the subsequent Judge module.
Using the retrieved chains as in-context reasoning exemplars, together with its domain knowledge, the LLM generates one or more plausible chains linking the target failure mode to intermediate physical effects and ultimately to sensors capable of observing those effects. The generated chains are not required to correspond to paths already stored in the CKG. The structured output of this stage is therefore a set of newly inferred diagnostic chains for each extracted failure mode.


\subsubsection{Final judgement}
\label{subsubsec:final_judgement}

Once all failure modes have been analysed, their inferred chains are passed to a separate LLM-based decision module, hereafter referred to as the Judge, together with the clarified question and the complete set of answer options. The Judge prompt is reported in Appendix~\ref{app:prompt_judge}.
The Judge then selects a single option identifier based on the generated diagnostic reasoning. 
The resulting inference procedure separates retrieval of stored graph knowledge, generation of failure-specific diagnostic chains, and matching of those chains to the available answer choices.
This separation assigns failure-specific analysis and final answer selection to distinct stages. The reasoning module receives candidate failure modes and retrieved exemplars, while the Judge receives the diagnostic question and answer choices to assess the relevance of the generated consequences. The answer choices can therefore influence reasoning indirectly through retrieval and failure-mode extraction.

\section{Dataset}
\label{sec:dataset}
The proposed approach is evaluated on FailureSensorIQ \cite{constantinides2025FailureSensorIQ}, one of the most complete and one of the few publicly available benchmarks specifically designed to assess the ability of LLMs to establish relationships between failure modes and monitoring sensors across different industrial assets. Developed by IBM Research, the dataset comprises 8,296 multiple-choice questions covering ten asset classes. The questions are generated through a template-based procedure using condition-monitoring knowledge from International Organization for Standardization (ISO) documentation, including ISO 17359. The benchmark addresses two complementary diagnostic tasks: FM2Sensor, which identifies the sensors relevant to a given failure mode, and Sensor-to-Failure-Mode (Sensor2FailureMode), which identifies possible failure modes associated with a sensor observation. Both tasks include positive and negative formulations, requiring the model to identify either the most relevant or the irrelevant option among the available answers.

FailureSensorIQ provides 2,667 Single-Correct Multiple-Choice Question Answering (SC-MCQA) items and 5,629 Multiple-Correct Multiple-Choice Question Answering (MC-MCQA) items. Consistently with the primary evaluation setting adopted in the original benchmark, this study considers only the single-correct subset. The assessment is further restricted to four asset classes for which suitable reference documentation was available to construct the corresponding CKGs: aero gas turbine, electric motor, fan, and power transformer. The resulting evaluation set comprises 1,314 questions. Table \ref{tab:dataset} reports the selected assets, the reference documents used for graph construction, and the number of questions considered for each asset.

An additional consideration worth mentioning concerning the interpretation of accuracy on FailureSensorIQ is raised in the original dataset study by Constantinides et al. \cite{constantinides2025FailureSensorIQ}. The benchmark is generated from failure-mode--sensor associations contained in ISO documentation and expressed through predefined question templates. These engineering relationships can be many-to-many: a failure mechanism may affect several measurable quantities and may therefore be observable through different sensing modalities. The single-correct formulation nevertheless requires one designated answer for each item, and expressions such as "most relevant sensor" consequently impose a ranking among relationships that may all be technically plausible to different degrees. Some sensor descriptions are also broad or technically underspecified, as illustrated by the "length measurement" example discussed in Appendix~\ref{app:clarify_example}.

This characteristic should be considered when interpreting the absolute accuracy values reported in the following section. The original FailureSensorIQ study evaluated five human participants on a balanced subset of 80 questions and reported a mean expert accuracy of approximately 60\%, with the best-performing expert reaching 66.19\%. Inter-rater agreement among the experts was moderate ($\kappa=0.462$) \cite{constantinides2025FailureSensorIQ}. These results do not establish a formal upper bound for the benchmark, since the human evaluation covers only a subset of the questions and expertise differs across asset classes. They do, however, indicate that disagreement and ambiguity are present even among human domain experts. Accordingly, no matter the methodology, the gap to 100 per cent accuracy should not be interpreted solely as model error, since a portion of the residual error may reflect ambiguity in the benchmark formulation and the non-unique nature of some failure--sensor relationships.

\begin{table}[htbp]
\centering
\caption{Industrial assets considered in the evaluation, reference
documentation used for CKG construction, and number of
single-correct FailureSensorIQ questions \cite{constantinides2025FailureSensorIQ}.}
\label{tab:dataset}
\renewcommand{\arraystretch}{1.08}
\setlength{\tabcolsep}{5pt}
\footnotesize

\begin{tabularx}{\textwidth}{p{2.5cm} X r}
\toprule
\textbf{Asset} &
\textbf{Reference documentation} &
\textbf{Questions} \\
\midrule

Aero gas turbine &
\textit{Gas Turbine Theory}, 7th ed.
\cite{saravanamuttoo2017gasturbine} &
336 \\

Electric motor &
\textit{Noninvasive Methods for Condition Monitoring
and Electrical Fault Diagnosis of Induction Motors}
\cite{sheikh2017noninvasive} &
234 \\

Fan &
\textit{Improving Fan System Performance:
A Sourcebook for Industry}
\cite{doe2003fan} &
200 \\

Power transformer &
\textit{Transformers: Basics, Maintenance, and Diagnostics}
\cite{usbr2005transformers} &
544 \\

\midrule
\textbf{Total} & & \textbf{1,314} \\
\bottomrule

\end{tabularx}
\end{table}

\section{Results}
\label{sec:results}

The evaluation was conducted on the 1,314 single-correct FailureSensorIQ questions retained for the four assets considered in this work. The section evaluates the proposed solution from the construction of the knowledge representation to the final diagnostic decision. Section~\ref{subsec:ckg_results} first examines the CKGs produced offline and their inspectability. Section~\ref{subsec:reasoning_analysis} then follows a worked reasoning example. Sections~\ref{subsec:diagnostic_accuracy}--\ref{subsec:asset_accuracy} assess diagnostic accuracy, the contribution of the inference stages, and consistency across assets. Section~\ref{subsec:no_ckg} removes the CKG entirely to isolate the contribution of the failure-to-observation reasoning structure. Together, these analyses assess the quality of the constructed knowledge, the benchmark performance of CoP, the role of its individual stages, and the contribution of its reasoning representation.

\subsection{CKG construction}
\label{subsec:ckg_results}

Table~\ref{tab:ckg_construction} summarises the outcome of the offline graph-construction procedure. The number of accepted chains varies substantially across the four reference documents. From 981 text chunks extracted from the aero gas turbine textbook, 55 complete diagnostic chains were retained, corresponding to a pass rate of 5.6\%. The fan source produced 25 chains from 152 chunks (16.4\%), while the electric-motor and power-transformer documents produced 14 from 38 chunks (36.8\%) and 135 from 333 chunks (40.5\%), respectively.

Figure~\ref{fig:ckg_structure} describes the structure of the resulting graphs. Most chains contain two or three relations, corresponding to one or two intermediate changes in physical quantities between the failure mode and the sensor. Longer chains, with up to five relations, are mainly extracted from the fan and aero gas turbine documents, whereas the electric-motor graph contains no chain longer than four relations. For all assets, intermediate physical-change entities form the largest group of nodes.

\begin{table}[htbp]
\centering
\caption{Outcome of the offline CKG construction for the four evaluated assets.}
\label{tab:ckg_construction}
\begin{tabular}{lrrrrr}
\toprule
\textbf{Asset} &
\textbf{Chunks} &
\textbf{Chains} &
\textbf{Pass rate} &
\textbf{Entities} &
\textbf{Relations} \\
\midrule
Aero gas turbine  & 981 & 55  & 5.6\%  & 209 & 164 \\
Electric motor    & 38  & 14  & 36.8\% & 44  & 35  \\
Fan               & 152 & 25  & 16.4\% & 99  & 79  \\
Power transformer & 333 & 135 & 40.5\% & 429 & 348 \\
\bottomrule
\end{tabular}

\vspace{4mm}
\includegraphics[width=0.85\textwidth]{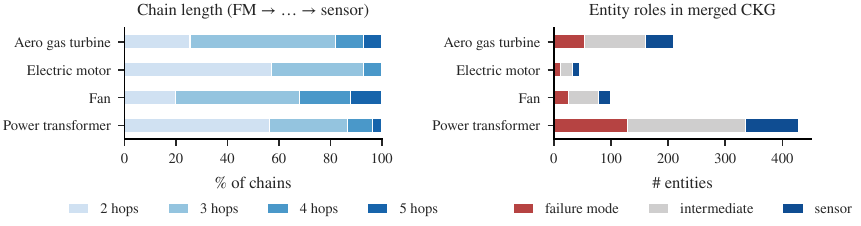}
\captionof{figure}{Structure of the four CKGs. Left: distribution of chain lengths, measured as the number of relations between the failure mode and the sensor. Right: number of failure-mode, intermediate physical-change, and sensor entities in each merged graph.}
\label{fig:ckg_structure}

\end{table}

The extraction rates in Table~\ref{tab:ckg_construction}
reflect the selective purpose of the CKG. Reference documents
contain extensive theoretical, historical, and design material,
while the extraction procedure retains passages supporting
complete failure-to-observation chains. The resulting graphs
contain between 14 and 135 accepted chains per asset, providing
a manageable body of diagnostic knowledge for independent
SME review. This compactness is a practical advantage of the
representation: each retained example exposes the physical
relationships used to connect an initiating condition to a
measurable effect.

\begin{figure}
    \centering
    \includegraphics[width=0.8\textwidth]{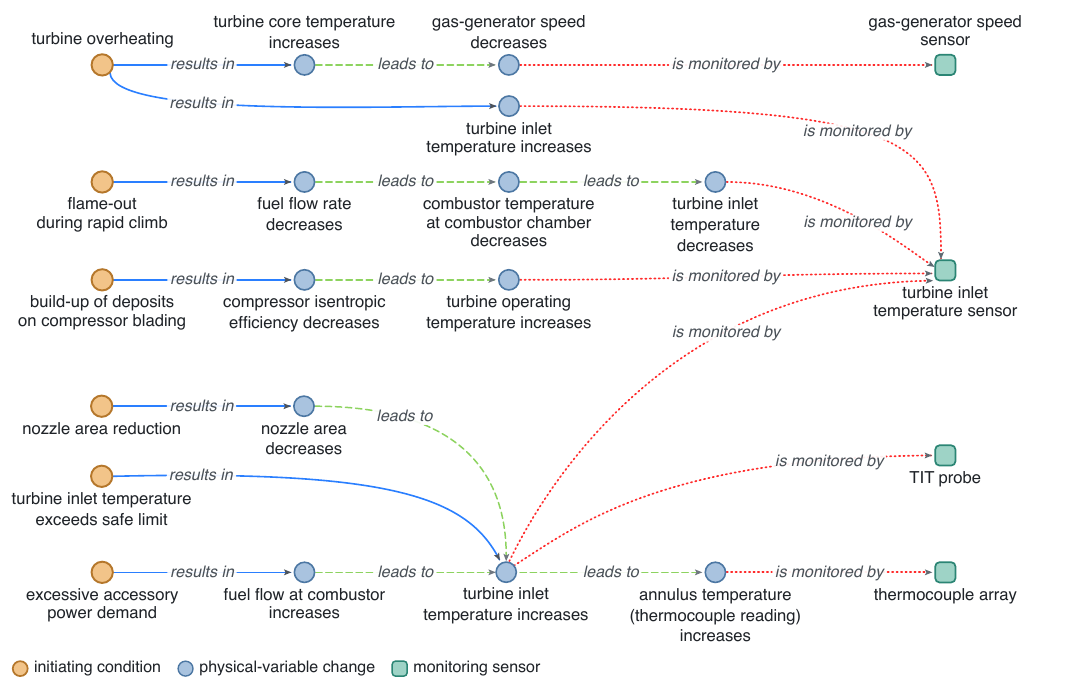}
    \caption{Excerpt from the aero gas turbine CKG. Orange nodes denote initiating conditions,
    blue nodes denote physical-variable changes, and green
    nodes denote monitoring sensors. The relations
    \textit{results in}, \textit{leads to}, and
    \textit{is monitored by} distinguish initial effects,
    physical propagation, and observation. The explicit chain
    structure supports step-by-step inspection by domain experts.}
    \label{fig:ckg_turbofan_excerpt}
\end{figure}

Figure~\ref{fig:ckg_turbofan_excerpt} illustrates a section of
the aero gas turbine CKG. The stored chains can contain different numbers of
intermediate physical changes, reflecting the amount of propagation supported
by the source material. The fixed relations distinguish the initial effect,
subsequent propagation, and sensor observation. This structure makes every
chain readable from the initiating failure to the monitored quantity and allows
an engineer to check its correctness step by step. The figure therefore also
illustrates the practical basis for the SME review described in
Section~\ref{subsubsec:graph_construction}.

During inference, retrieved chains serve as engineering
knowledge and few-shot demonstrations of diagnostic reasoning.
Their role is to provide complete examples of how physical
effects connect an initiating condition to observable
quantities. CoP therefore uses a selective set of reviewed
diagnostic exemplars to guide the analysis of candidate failure
modes, including those absent from the retrieved paths.

\subsection{Analysis of the reasoning process}
\label{subsec:reasoning_analysis}

\begin{figure}
    \centering
    \includegraphics[width=\textwidth]{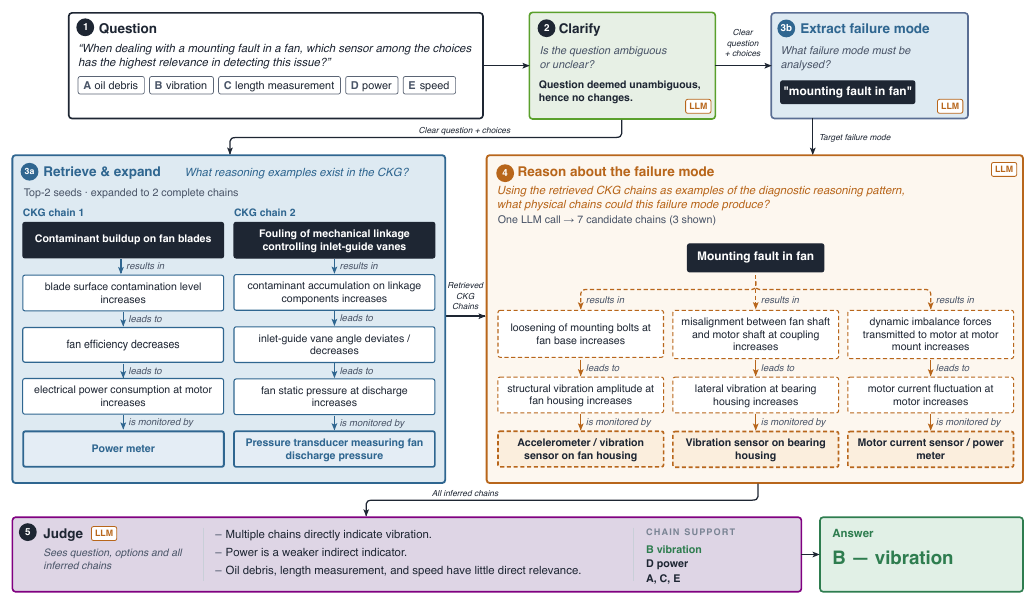}
    \caption{Example of the CoP inference process for a fan mounting fault. The input question is first assessed by the clarification module and is left unchanged because no ambiguity is detected. The target failure mode is extracted directly from the clarified question, while the CKG retrieval stage independently recovers complete stored chains associated with the most relevant graph entities. These paths provide examples of the expected failure--physical effect--sensor reasoning structure. The reasoning stage then generates new plausible chains for the mounting fault, including paths leading to structural and lateral vibration measurements. The Judge receives the question, answer choices, and all inferred chains, and selects vibration as the final answer.}
    \label{fig:cop_reasoning_example}
\end{figure}

The illustrative example in Figure~\ref{fig:cop_reasoning_example} illustrates one case of how the CoP inference procedure operates in practice. A second example reporting the complete end-to-end output of every online stage, including retrieval seeds, expanded CKG chains, inferred chains, and Judge reasoning, is provided in Appendix~\ref{app:full_trace}. The question concerns a mounting fault in a fan and asks which of five candidate measurements has the highest diagnostic relevance. The clarification module identifies the item as sufficiently clear and leaves the question unchanged. The failure-mode extraction stage identifies \textit{mounting fault in fan} as the condition to be analysed.

In parallel, the retrieval stage searches the fan CKG using the clarified question and answer options. The two recovered chains concern contaminant buildup on fan blades and fouling of the inlet-guide-vane linkage. Their physical paths terminate at a power meter and a pressure transducer, respectively. Neither retrieved chain contains the mounting fault appearing in the question or directly points to vibration, which is the correct answer.

The reasoning stage then uses these stored chains as examples of the expected failure-to-observation structure and applies the same reasoning pattern to the extracted mounting fault. Among the generated chains, the model links mounting faults to loose mounting bolts, shaft misalignment, and transmitted imbalance forces. Two of these chains terminate in structural or lateral vibration measurements, while another terminates in motor-current or power measurement. The Judge subsequently compares the complete set of inferred chains with the available answer choices and selects \textit{vibration}, corresponding to the gold label.

This example is useful because the retrieved CKG paths provide limited topical overlap with the target failure mode. Their contribution lies in supplying an asset-specific representation of how a failure is connected to physical effects and observable quantities. The new mounting-fault chains are generated during the reasoning stage and are therefore distinct from the paths stored in the CKG. Figure~\ref{fig:cop_reasoning_example} consequently shows the intended interaction between graph retrieval and LLM inference: the graph provides structured diagnostic examples, while the model extends the same structure to the failure mode under analysis.

The trace should be interpreted as an explanatory example of the inference process, while the quantitative results reported in the previous sections provide the evidence for the observed performance improvements. The clarification and decision stages further improve overall accuracy. The results are based on one run for each model with zero temperature.

\subsection{Diagnostic accuracy}
\label{subsec:diagnostic_accuracy}

The overall accuracy obtained by the three approaches is reported in Table~\ref{tab:main_results}. CoP achieves the highest accuracy for all three models. With Claude Opus 4.7, accuracy increases from 51.98\% with direct prompting and 50.61\% with RAG to 68.80\% with CoP. The same trend is observed for the smaller models: Llama 3.1 8B increases from 30.75\% to 38.51\%, while Qwen 2.5 7B increases from 30.52\% to 50.15\%. Overall, the gain over direct prompting ranges from 7.8 to 19.6 percentage points, while the gain over conventional RAG ranges from 11.7 to 18.2 percentage points. The absolute values should be interpreted in light of the benchmark characteristics discussed in Section~\ref{sec:dataset}.

\begin{table}[htbp]
\centering
\caption{Overall diagnostic accuracy on the 1,314 single-correct FailureSensorIQ questions.}
\label{tab:main_results}
\begin{tabular}{lccc}
\toprule
\textbf{Model} & \textbf{Direct} & \textbf{RAG} & \textbf{CoP} \\
\midrule
Claude Opus 4.7 & 51.98 & 50.61 & \textbf{68.80} \\
Llama 3.1 8B   & 30.75 & 26.86 & \textbf{38.51} \\
Qwen 2.5 7B    & 30.52 & 35.24 & \textbf{50.15} \\
\bottomrule
\end{tabular}
\end{table}

The comparison with RAG is particularly relevant because both approaches rely on the same reference documentation and the same initial chunking procedure. The difference therefore lies in the representation and use of the retrieved information. RAG provides the model with semantically similar text passages, while CoP retrieves complete failure-to-observation chains from the CKG and uses them as reasoning examples. The results indicate that the structure imposed on the engineering knowledge has a substantial influence on the final diagnostic performance. This effect is visible in all three models and is especially pronounced for Qwen 2.5 7B, for which CoP improves accuracy by almost 15 percentage points over RAG.

The RAG results also show that the availability of additional textual context does not translate consistently into higher accuracy. This result is aligned with the conclusion in \cite{constantinides2025FailureSensorIQ}, where a RAG pipeline is implemented. Claude Opus 4.7 and Llama 3.1 8B both obtain slightly lower performance with RAG than with direct prompting, while Qwen 2.5 7B benefits from retrieved context. This variability suggests that the contribution of retrieval depends on how effectively the model can reconstruct the relevant diagnostic relationship from the retrieved passages. CoP reduces this burden by retrieving information already organised according to the failure--physical effect--sensor structure used during reasoning.

\subsection{Contribution of the inference stages}
\label{subsec:ablation}

The ablation study in Table~\ref{tab:ablation_results} examines how the additional inference stages affect the performance obtained from the CKG. All configurations use the same CKG, the same embedding-based seed selection ($k=2$), and the same path expansion. The configurations differ in the LLM calls surrounding retrieval:

\begin{itemize}
    \item \textbf{CoP w/o clarify + Judge.} The original question and options are used directly for retrieval. No clarification call, failure-mode extraction call, failure-specific reasoning call, or separate Judge is used. One LLM call receives the original question, the answer options, the retrieved CKG chains, and the chain pattern, and jointly produces the diagnostic reasoning and the selected option.

    \item \textbf{CoP w/o Judge.} The clarification call is added before retrieval. The clarified question and options are then passed to the same single answer-generation call used in the previous configuration. Failure-mode extraction and separate failure-specific reasoning are still absent, and reasoning and option selection remain within one LLM call. The comparison with the previous configuration therefore isolates the effect of clarification.

    \item \textbf{Full CoP.} The clarification stage is retained, and the single answer-generation call is replaced by separate failure-mode extraction, one reasoning call for each extracted failure mode, and the final Judge call described in Section~\ref{subsec:online_inference}. The reasoning calls do not receive the answer choices directly. The comparison with \textit{CoP w/o Judge} therefore evaluates the combined effect of failure-mode extraction, separate failure-specific reasoning, and final answer selection through the Judge.
\end{itemize}

\begin{table}[htbp]
\centering
\caption{Ablation study of the CoP inference procedure.}
\label{tab:ablation_results}
\begin{tabular}{lccc}
\toprule
\textbf{Model} &
\textbf{w/o clarify + Judge} &
\textbf{w/o Judge} &
\textbf{Full CoP} \\
\midrule
Claude Opus 4.7 & 66.36 & 67.81 & \textbf{68.80} \\
Llama 3.1 8B   & 35.16 & 36.61 & \textbf{38.51} \\
Qwen 2.5 7B    & 43.53 & 48.55 & \textbf{50.15} \\
\bottomrule
\end{tabular}

\vspace{4mm}
\includegraphics[width=0.8\textwidth]{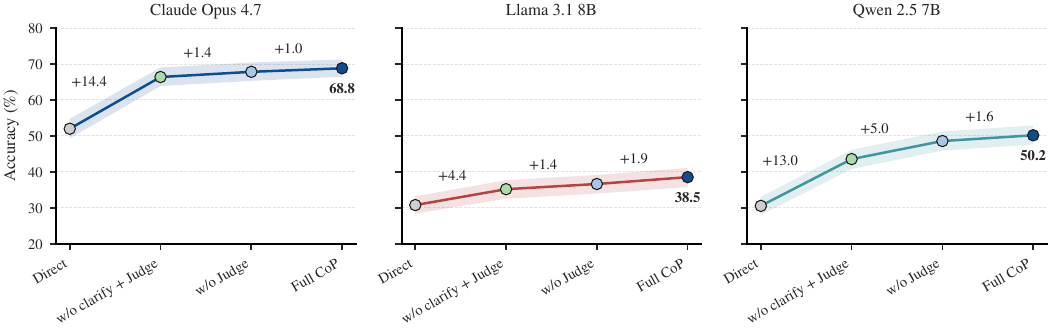}
\captionof{figure}{Accuracy gain contributed by each inference stage, starting from direct prompting. Each point adds one component to the previous configuration. Labels report the gain in percentage points.}
\label{fig:staircase}

\end{table}

The configuration combining CKG retrieval with a single answer-generation call already reaches 66.36\%, 35.16\%, and 43.53\% for the three models, respectively. These values account for a large part of the improvement observed with the complete method and support the role of the constrained graph representation itself. Clarification adds 1.45 percentage points for Claude Opus 4.7 and Llama 3.1 8B, and 5.02 points for Qwen 2.5 7B. The full staged inference procedure adds a further 0.99, 1.90, and 1.60 percentage points, respectively. Figure~\ref{fig:staircase} shows the same results as a cumulative sequence starting from direct prompting. For all three models, the largest single increase is obtained when CKG retrieval is introduced, while clarification and the Judge provide smaller additional gains.

\subsection{Accuracy across asset classes}
\label{subsec:asset_accuracy}

Table~\ref{tab:asset_results} reports diagnostic accuracy separately for each asset class, including the intermediate inference configurations. Full CoP outperforms direct prompting and conventional RAG in all twelve asset--model combinations. The gains range from 4.28 to 21.14 percentage points over direct prompting and from 8.93 to 20.51 percentage points over RAG. The aggregate improvement is therefore observed across all four evaluated asset classes. Figure~\ref{fig:asset_radar} reports the same comparison for each asset. With Claude Opus 4.7 and Qwen 2.5 7B, CoP obtains the highest accuracy on all four assets. With Llama 3.1 8B, the margins are smaller. CoP is most clearly ahead on the electric motor and the fan, while the three approaches remain close on the aero gas turbine and the power transformer.

\begin{table}[htbp]
\centering
\caption{Diagnostic accuracy (\%) by asset class, model, and inference configuration. CKG denotes graph retrieval with direct answer generation; CoP denotes the complete inference procedure, including clarification, separate failure-specific reasoning, and final judgement. The number of evaluated questions is reported once for each asset. Bold values indicate the highest accuracy within each row.}
\label{tab:asset_results}
\small
\setlength{\tabcolsep}{5pt}
\begin{tabular}{llrrrr}
\toprule
\textbf{Asset ($N$)} & \textbf{Model} & \textbf{Direct} & \textbf{RAG} & \textbf{CKG} & \textbf{CoP} \\
\midrule
Aero gas turbine (336) & Claude Opus 4.7 & 43.45 & 44.05 & 59.82 & \textbf{61.01} \\
 & Llama 3.1 8B & 29.17 & 30.06 & 34.52 & \textbf{38.99} \\
 & Qwen 2.5 7B & 27.68 & 31.85 & 37.80 & \textbf{49.40} \\
\addlinespace
Electric motor (234) & Claude Opus 4.7 & 66.67 & 54.70 & 66.67 & \textbf{75.21} \\
 & Llama 3.1 8B & 36.32 & 24.36 & \textbf{40.60} & \textbf{40.60} \\
 & Qwen 2.5 7B & 29.91 & 35.47 & 44.44 & \textbf{50.43} \\
\addlinespace
Fan (200) & Claude Opus 4.7 & 51.50 & 51.00 & 63.00 & \textbf{65.00} \\
 & Llama 3.1 8B & 37.50 & 27.00 & 43.50 & \textbf{45.50} \\
 & Qwen 2.5 7B & 38.00 & 46.50 & 53.00 & \textbf{60.00} \\
\addlinespace
Power transformer (544) & Claude Opus 4.7 & 51.10 & 52.76 & 71.51 & \textbf{72.24} \\
 & Llama 3.1 8B & 26.84 & 25.92 & 30.15 & \textbf{34.74} \\
 & Qwen 2.5 7B & 29.78 & 33.09 & 43.01 & \textbf{46.88} \\
\addlinespace
\bottomrule
\end{tabular}

\vspace{4mm}
\includegraphics[width=\textwidth]{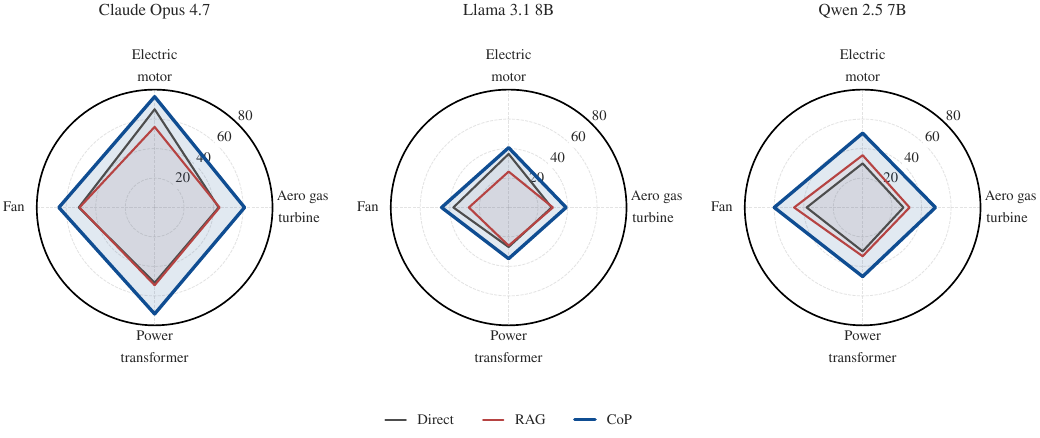}
\captionof{figure}{Diagnostic accuracy (\%) per asset for direct prompting, RAG, and CoP with the three evaluated LLMs.}
\label{fig:asset_radar}

\end{table}

The intermediate stages show greater variation. For Llama 3.1 8B, clarification reduces accuracy from 34.52\% to 32.44\% on the aero gas turbine and from 40.60\% to 40.17\% on the electric motor. Adding separate failure-specific reasoning and final judgement reduces fan accuracy from 46.50\% to 45.50\% for Llama, and power-transformer accuracy from 72.98\% to 72.24\% for Claude and from 47.98\% to 46.88\% for Qwen. It leaves Qwen's fan accuracy unchanged at 60.00\%. These results qualify the aggregate ablation: the additional stages improve overall accuracy, but their benefit depends on the asset and underlying model.

These results also indicate that the contribution of the individual stages depends on the underlying model. Question clarification has a limited effect for Claude Opus 4.7 and Llama 3.1 8B, while its effect is more marked for Qwen 2.5 7B. The Judge provides a smaller increase in aggregate accuracy across all three models, although the asset-level results show exceptions.

\subsection{Accuracy by question type}
\label{subsec:question_type}
 
\begin{figure}
    \centering
    \includegraphics[width=\textwidth]{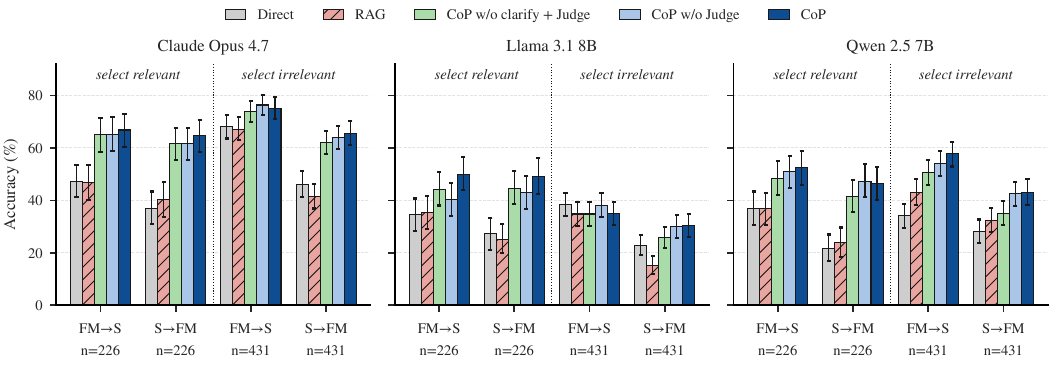}
    \caption{Accuracy (\%) by question type for direct prompting, RAG, and the three CoP configurations. (FM$\rightarrow$S: failure mode to sensor; S$\rightarrow$FM: sensor to failure mode; $n$: number of questions of each type.)}
    \label{fig:question_type}
\end{figure}

FailureSensorIQ contains four question types, obtained by combining the reasoning direction (from failure mode to sensor, FM$\rightarrow$S, or from sensor to failure mode, S$\rightarrow$FM) with the polarity of the question (selecting the relevant or the irrelevant option). Figure~\ref{fig:question_type} reports the accuracy of all configurations for each type.
 
With Claude Opus 4.7, the largest improvements occur for questions that require selecting the relevant option, where CoP increases accuracy by approximately 20 to 30 percentage points in both directions, and for S$\rightarrow$FM questions with negative polarity. FM$\rightarrow$S questions with negative polarity are already answered comparatively well by direct prompting and show a smaller gain. With Qwen 2.5 7B, CoP improves accuracy for all four question types. With Llama 3.1 8B, CoP improves accuracy for the two positive-polarity types and for S$\rightarrow$FM questions with negative polarity, but not for FM$\rightarrow$S questions with negative polarity, where direct prompting remains slightly better. S$\rightarrow$FM questions with negative polarity are the most difficult type for this model across all approaches. Overall, the improvement obtained with CoP is not limited to a single question formulation, although its magnitude depends on both the question type and the model.

\subsection{Reasoning structure without the CKG}
\label{subsec:no_ckg}

The previous experiments evaluate CoP with both the constrained failure-to-observation reasoning structure and the asset-specific knowledge stored in the CKG. To examine the contribution of the reasoning structure independently of graph retrieval, we also evaluate a variant without CKG support using the multi-agent architecture proposed by Gong et al.~\cite{gong2026sensorreasoning}. This study was selected because it is one of the few approaches evaluated on FailureSensorIQ for which the agent roles, routing logic, and overall decision procedure are described in sufficient detail to support an independent reimplementation.

The architecture described by Gong et al. comprises a Guide Agent, a Discriminator Agent, a Simple Answerer Agent, one Option Advocate Agent for each answer choice, and a Judge Agent. The Guide produces a reasoning blueprint, while the Discriminator classifies each question as \textit{SIMPLE} or \textit{COMPLEX}. SIMPLE questions are routed directly to the Simple Answerer. COMPLEX questions are processed by the Option Advocates, after which the Judge selects the final answer from the generated arguments.

Since the original study does not provide a public code repository or the complete prompts used for the individual agents, the framework was reconstructed from the methodological description, Algorithm~1, and the agent responsibilities reported in the paper. The resulting implementation follows the published agent structure and SIMPLE/COMPLEX routing procedure.

Within this reimplementation, the comparison is designed to isolate the form of the reasoning exchanged between the Option Advocates and the Judge. In the Gong configuration, each advocate produces a free-text argument supporting its assigned answer option. In the CoP w/o CKG configuration, the same advocate is required to express its argument as a diagnostic chain of the form

\[
\text{failure mode}
\rightarrow
\text{physical change}_1
\rightarrow \cdots \rightarrow
\text{sensor}.
\]

The chain is enforced through structured output and then passed to the Judge in place of the free-text argument. The remaining architecture and routing procedure are kept identical within our reimplementation. No CKG, graph retrieval, or reference documentation is used in either configuration. The diagnostic chains are therefore generated entirely from the model's internal knowledge.

Since this experiment does not require asset-specific graph construction, it is conducted on all ten asset classes and on the complete set of 2,667 single-correct FailureSensorIQ questions. Both configurations are evaluated with Llama 3.1 8B and Qwen 2.5 7B. In the original Gong et al. study, the Guide, Discriminator, and Judge use GPT-4o-mini while the evaluated open-source model is used for the remaining roles. In our comparison, all agents are instantiated with the same model. This removes the contribution of a stronger orchestration model and allows the effect of the reasoning representation to be assessed under a homogeneous-model setting.

The results are reported in Table~\ref{tab:no_ckg_results}. For Llama 3.1 8B, replacing free-text advocacy with the constrained failure-to-observation chain increases overall accuracy from 43.34\% to 45.14\%. For Qwen 2.5 7B, accuracy increases from 49.34\% to 53.28\%. Across the individual asset classes, CoP w/o CKG improves performance in 16 of the 20 asset--model combinations, matches the Gong configuration in one case, and performs lower in three.

\begin{table}[htbp]
\centering
\caption{Accuracy (\%) on the 2,667 single-correct FailureSensorIQ questions without graph retrieval. Gong et al. denotes the reimplemented multi-agent framework of \cite{gong2026sensorreasoning} with all agents instantiated using the same model.}
\label{tab:no_ckg_results}
\renewcommand{\arraystretch}{1.1}
\setlength{\tabcolsep}{4pt}
\footnotesize
\begin{tabular}{llccccccccccc}
\toprule
\textbf{Model} & \textbf{Method} & \textbf{AGT} & \textbf{CMP} & \textbf{EG} & \textbf{EM} & \textbf{FAN} & \textbf{IGT} & \textbf{PT} & \textbf{PMP} & \textbf{RICE} & \textbf{ST} & \textbf{Overall} \\
\midrule
\multirow{2}{*}{Llama 3.1 8B}
 & Gong et al.  & \textbf{41.96} & 40.45 & 41.88 & 44.44 & 46.00 & \textbf{52.50} & 36.95 & \textbf{51.97} & 47.92 & 38.01 & 43.34 \\
 & CoP w/o CKG  & 41.67 & \textbf{45.91} & \textbf{45.30} & \textbf{46.58} & \textbf{50.50} & 52.08 & \textbf{38.79} & \textbf{51.97} & \textbf{48.81} & \textbf{39.77} & \textbf{45.14} \\
\midrule
\multirow{2}{*}{Qwen 2.5 7B}
 & Gong et al.  & 41.67 & 47.73 & 52.99 & \textbf{53.85} & 54.00 & 55.00 & 43.38 & 52.63 & 56.85 & 43.27 & 49.34 \\
 & CoP w/o CKG  & \textbf{45.54} & \textbf{51.82} & \textbf{58.55} & 50.00 & \textbf{57.50} & \textbf{57.50} & \textbf{50.74} & \textbf{53.95} & \textbf{62.20} & \textbf{46.78} & \textbf{53.28} \\
\midrule
\multicolumn{2}{l}{Questions} & 336 & 220 & 234 & 234 & 200 & 240 & 544 & 152 & 336 & 171 & 2,667 \\
\bottomrule
\end{tabular}
\end{table}

The improvement is observed without any contribution from graph-derived knowledge or retrieved engineering documentation. The difference between the two configurations lies in the representation of the intermediate diagnostic argument presented to the Judge. Constraining this argument to an explicit failure-to-observation chain therefore improves the final decision even when the chain is generated entirely from the model's internal knowledge.

This result is relevant for the interpretation of the complete CoP framework. The CKG contributes asset-specific engineering knowledge and retrieved examples of diagnostic propagation, while the chain representation determines how the intermediate reasoning is expressed. The experiment in this subsection shows that the latter remains beneficial when the CKG is removed. The gains obtained with the full framework can therefore be associated with two distinct elements: the engineering knowledge supplied by the graph and the explicit physical structure imposed on the reasoning process.

\section{Conclusions and Future Work}
\label{sec:conclusions}

LLMs offer a practical means of supporting knowledge-intensive engineering tasks, provided that the information and reasoning structure required by the task are made explicit. This study introduced CoP for industrial fault diagnosis, translating the way an engineer traces a failure through its physical consequences into a constrained failure-to-observation representation. Asset-specific reference documentation is converted offline into a CKG, and complete graph paths are retrieved during inference as engineering knowledge and reasoning examples for the generation of failure-specific diagnostic chains.
CoP addresses a complementary diagnostic problem to conventional data-driven fault classification: reasoning over engineering knowledge to establish the relationships between failure mechanisms, their physical effects, and the sensing modalities through which those effects can be observed.

On 1,314 single-correct FailureSensorIQ questions, CoP improved accuracy across all three evaluated LLMs, with gains of 7.8--19.6 percentage points over direct prompting and 11.7--18.2 points over conventional RAG. The ablation analysis showed that the CKG provides a substantial part of this improvement, while the additional inference stages provide further gains. The comparison without graph retrieval provides a second result: imposing the same failure-to-observation structure within the reproducible framework of Gong et al. improves overall accuracy for both tested models. This supports the use of physically ordered diagnostic chains as a reasoning representation independently of the asset-specific knowledge supplied by the CKG.

Future work will extend the evaluation to real industrial datasets combining sensor measurements, maintenance records, and naturally occurring diagnostic requests. Further work should also test alternative reasoning architectures and richer graph constraints, including component topology, operating-state dependencies, and admissible propagation paths, and develop confidence or abstention mechanisms for cases in which the available engineering knowledge does not support a sufficiently reliable diagnostic conclusion.


\section*{CRediT Authorship Contribution Statement}
\textbf{Wenqi Zhou:} Conceptualization, Methodology, Software, Validation, Investigation, Visualization, Writing - Original Draft, Writing - Review \& Editing
\textbf{Leonardo Baldo:} Methodology, Visualization, Writing - Original Draft, Writing - Review \& Editing
\textbf{Kai Goebel:} Conceptualization, Methodology, Writing - Review \& Editing
\textbf{Manuel Arias Chao:} Conceptualization, Methodology, Validation, Writing - Review \& Editing, Supervision, Funding acquisition

\section*{Declaration of Competing Interest}
The authors declare that they have no known competing financial interests or personal relationships that could have appeared to influence the work reported in this paper.

\section*{Code and Data Availability Statement}
Data sharing is not applicable to this article as no new data were created or generated. The FailureSensorIQ dataset employed in this study is publicly available \cite{constantinides2025FailureSensorIQ}. The code will be made available upon request.

\section*{Declaration of generative AI and AI-assisted technologies in the manuscript preparation process}
During the preparation of this work, the authors used ChatGPT and Claude to assist with language editing, image editing, improving readability, and refining the presentation of the manuscript. After using these tools, the authors reviewed and edited the content as needed and take full responsibility for the content of the published article.

\bibliographystyle{unsrt}
\bibliography{References}

\appendix

\section{Prompts}
\label{app:prompts}

The online CoP pipeline uses four LLM calls: clarification, failure-mode extraction, failure-mode reasoning, and final judging. Each module uses a dedicated system instruction and a structured user prompt. The templates are reported below. Terms enclosed in angle brackets indicate values populated at inference time.

\subsection{Offline CKG Construction}
\label{app:prompt_offline}

The offline CKG construction uses one LLM call for each text unit extracted from the reference documentation. The model is asked to identify one complete failure-to-observation chain grounded in the current text unit. The output is constrained to a structured schema containing one failure mode, one to five changes in physical quantities, and one monitoring sensor.

\begin{promptbox}{Diagnostic-Chain Extraction System Prompt}

You extract ONE failure-mode $\rightarrow$ sensor causal chain for asset: \texttt{<Asset>}.

\medskip
Describe ONE coherent monitoring scenario on a \texttt{<Asset>} as a chain:

\medskip
\texttt{failure\_mode $\rightarrow$ variable\_1 changes $\rightarrow$ variable\_2 changes $\rightarrow$ ... $\rightarrow$ sensor}

\medskip
Fields:

\smallskip
\noindent$\bullet$ \texttt{failure\_mode}: named fault / degradation / abnormal condition of the \texttt{<Asset>} (or a named subcomponent).

\noindent$\bullet$ \texttt{steps}: ordered list of 1--5 physical-variable changes. Each step contains:

\hspace{4mm}-- \texttt{variable}: the physical quantity that moves;

\hspace{4mm}-- \texttt{location}: where on the \texttt{<Asset>} it is taken; empty if the CHUNK does not say;

\hspace{4mm}-- \texttt{direction}: ``increases'' or ``decreases''.

\noindent$\bullet$ \texttt{steps[0]} is what the failure mode drives directly; each later step is caused by the one before it; the LAST step is what the sensor reads.

\noindent$\bullet$ \texttt{sensor}: instrument that reads the LAST step's variable.

\medskip
If the CHUNK does not ground the \texttt{failure\_mode}, at least one step, AND the sensor for the SAME scenario --- or is only general theory / history / equations / design with no failure-to-monitoring link --- leave \texttt{failure\_mode} and \texttt{sensor} empty and \texttt{steps=[]}.

\medskip
Rules:

\smallskip
\noindent$\bullet$ Use ONLY facts grounded in the CHUNK. Do NOT add physics hops the CHUNK does not state, and do NOT pad the chain to make it longer. A 1-step chain is correct when that is all the CHUNK supports.

\noindent$\bullet$ Every step must be a physical variable that goes up or down --- not a fault, not an instrument, not a maintenance action.

\noindent$\bullet$ Do not repeat the same variable+location+direction twice in the chain.

\noindent$\bullet$ Prefer the dominant failure scenario when several appear.

\noindent$\bullet$ Stay specific to \texttt{<Asset>}; do not import failure modes from other asset classes.

\noindent$\bullet$ Keep each field a short phrase; no markdown.

\end{promptbox}

\begin{promptbox}{Diagnostic-Chain Extraction User Prompt Template}

CHUNK:

\texttt{<Reference Text Unit>}

\end{promptbox}

The structured output contains the fields \texttt{failure\_mode}, \texttt{steps}, and \texttt{sensor}. Each element of \texttt{steps} contains a \texttt{variable}, an optional \texttt{location}, and a \texttt{direction} restricted to \texttt{increases} or \texttt{decreases}. Empty outputs are interpreted as rejected text units and do not contribute to the CKG.

\subsection{Clarification}
\label{app:prompt_clarification}

The clarification module checks whether the original multiple-choice question or its options require rewording before the diagnostic process starts. The answer choices are preserved in number, order, and meaning.

\begin{promptbox}{Clarification System Prompt}

You clarify failure-mode / sensor multiple-choice questions before they are answered.

\medskip
Rules:

\smallskip
\noindent$\bullet$ First decide \texttt{needs\_clarification}.

\noindent$\bullet$ If already clear: set \texttt{needs\_clarification=false} and leave \texttt{clarified\_question} empty and \texttt{clarified\_options} as \texttt{[]}. Do not rewrite anything.

\noindent$\bullet$ If clarification is needed: set \texttt{needs\_clarification=true}, rewrite \texttt{clarified\_question}, and return \texttt{clarified\_options} with the same number/order of options.

\noindent$\bullet$ Preserve the original meaning, asset, failure event, and which sensor each option refers to.

\noindent$\bullet$ Do not add, drop, merge, or reorder options when rewriting.

\noindent$\bullet$ Do not pick an answer. Do not invent new failure modes or sensors not implied by the text.

\noindent$\bullet$ Make vague wording concrete without changing substance.

\end{promptbox}

\begin{promptbox}{Clarification User Prompt Template}

Question: \texttt{<Question>}

\medskip
Options:

\texttt{<Option ID 1> <Option 1>}\\
\texttt{<Option ID 2> <Option 2>}\\
\texttt{...}

\medskip
There are exactly \texttt{<N>} options with ids \texttt{<Option IDs>}.

If \texttt{needs\_clarification=true}, return \texttt{clarified\_options} with the same length and order; if false, leave \texttt{clarified\_question} empty and \texttt{clarified\_options=[]}.

\end{promptbox}

The structured output contains \texttt{needs\_clarification}, \texttt{clarified\_question}, \texttt{clarified\_options}, and a short note describing the ambiguity when present.

\subsection{Failure-Mode Extraction}
\label{app:prompt_failure_extraction}

Failure modes are extracted from the clarified question and answer options. This step is independent of the CKG.

\begin{promptbox}{Failure-Mode Extraction System Prompt}

You are a failure analysis expert. Extract all failure modes mentioned or implied in the given question and answer options.

A failure mode is a way in which a component, system, or process can fail to perform its intended function.

\medskip
Look for:

\smallskip
\noindent$\bullet$ Direct mentions of failure modes, e.g. ``bearing wear'' or ``shaft misalignment''.

\noindent$\bullet$ Implied failures described by symptoms or effects.

\noindent$\bullet$ Failure conditions mentioned in context.

\medskip
Extract the failure mode names as they naturally appear in the text. Be precise and avoid over-interpretation.

\end{promptbox}

\begin{promptbox}{Failure-Mode Extraction User Prompt Template}

Question: \texttt{<Clarified Question>}

\medskip
Options:

\texttt{<Option ID 1> <Option 1>}\\
\texttt{<Option ID 2> <Option 2>}\\
\texttt{...}

\medskip
Extract all failure modes mentioned or implied in the above text.

A failure mode is a way a component or system can fail.

Look for direct mentions, descriptions of failures, or failure conditions.

\end{promptbox}

The output is a list of extracted failure modes. Each failure mode is subsequently analysed independently.

\subsection{Failure-Mode Reasoning}
\label{app:prompt_reasoning}

For each extracted failure mode, the model receives the retrieved CKG chains and the corresponding chain structure. The retrieved paths provide examples of failure-to-sensor reasoning and are used to guide the generation of plausible diagnostic chains for the target failure mode.

\begin{promptbox}{Failure-Mode Reasoning System Prompt}

You are a failure analysis expert. Given a specific failure mode and knowledge graph examples, identify what diagnostic chains this failure mode could produce.

Use the knowledge graph examples as patterns to understand how failure modes typically lead to signatures and then to sensors.

Do NOT try to answer any specific question. Analyze the failure mode itself and list the possible diagnostic chains it could lead to.

\end{promptbox}

\begin{promptbox}{Failure-Mode Reasoning User Prompt Template}

Asset: \texttt{<Asset>}

\medskip
Failure mode to analyze: \texttt{<Failure Mode>}

\medskip
Knowledge graph examples showing diagnostic chain patterns for \texttt{<Asset>}:

\texttt{<Retrieved CKG Chains>}

\medskip
Chain pattern:

\texttt{<Failure Mode> -- results in --> <Signature> -- leads to --> <Signature> -- is monitored by --> <Sensor>}

\medskip
Based on these examples and your domain knowledge about \texttt{<Asset>} systems, analyze what diagnostic chains the failure mode \texttt{<Failure Mode>} could produce.

What signatures might this failure mode cause?

What sensors could detect those signatures?

Do not reference any specific question or try to pick an answer. Analyze the failure mode itself in the context of \texttt{<Asset>} systems.

\end{promptbox}

The structured output contains the list of possible diagnostic chains inferred for the target failure mode.

\subsection{Judge}
\label{app:prompt_judge}

The Judge receives the clarified question, the answer choices, the task polarity, and the reasoning generated for all extracted failure modes. It compares the inferred chains with the available choices and returns one option.

\begin{promptbox}{Judge System Prompt}

You are the Judge Agent for failure-mode / sensor multiple-choice questions.

You will receive independent reasoning about failure modes extracted from the question/options, along with the original question and options.

Use the failure mode reasonings to understand what diagnostic chains are possible, then match those to the options to select the best answer.

Follow the task polarity exactly: for MOST relevant select the strongest match; for LEAST relevant select the weakest or most contradicted match.

\end{promptbox}

\begin{promptbox}{Judge User Prompt Template}

Question: \texttt{<Clarified Question>}

\medskip
Options:

\texttt{<Option ID 1> <Option 1>}\\
\texttt{<Option ID 2> <Option 2>}\\
\texttt{...}

\medskip
\texttt{<Task Polarity Instruction>}

\medskip
Independent failure mode analysis:

\medskip
Failure Mode: \texttt{<Failure Mode 1>}

Possible chains: \texttt{<Possible Chains 1>}

\medskip
\texttt{...}

\medskip
Based on the failure mode reasonings above, determine which option best matches the diagnostic chains.

Select exactly one option from \texttt{<Candidate Option IDs>}.

\end{promptbox}

For positive questions, the task instruction asks for the strongest and most direct diagnostic relationship. For negative questions, it asks for the weakest, most indirect, or most contradicted relationship. The Judge returns a comparative reasoning field and one answer identifier.


\section{Worked Example of Question Clarification}
\label{app:clarify_example}

The ablation study in Table~\ref{tab:ablation_results} shows that the clarification stage increases accuracy for all three models. The trace in Figure~\ref{fig:cop_reasoning_example} illustrates the case in which the question is already clear and is passed through unchanged. This appendix reports the complementary case, in which the clarification module identifies an ambiguity, rewrites the question, and changes the final answer.

The example is taken from the fan subset of FailureSensorIQ and was obtained with Claude Opus 4.7.

\subsection{Original question}
\label{app:clarify_original}

\begin{promptbox}{Input Question (Fan, Sensor2FailureMode, negative)}

\textbf{Stem.}
Which failure event is irrelevant for fan if there is an abnormal reading from the sensor length measurement?

\medskip

\begin{tabularx}{\linewidth}{
    >{\bfseries}p{7mm}
    X
    >{\raggedleft\arraybackslash}p{15mm}
}
A & damaged oil seals & \\[0.8mm]
B & bearing wear & \\[0.8mm]
C & bearing damage & \\[0.8mm]
\textcolor{answerblue}{D} &
\textbf{\textcolor{answerblue}{rotor fouled}} &
\textbf{\textcolor{answerblue}{Gold}} \\[0.8mm]
E & misalignment &
\end{tabularx}

\end{promptbox}

The question contains two sources of ambiguity. First, the sensor is described as a \textit{length measurement}, which does not specify the physical quantity being measured. Second, the negative formulation (\textit{irrelevant \ldots\ if there is an abnormal reading}) can be read either as ``which failure is unimportant in general'' or as ``which failure cannot be indicated by this sensor''. Only the second reading corresponds to the diagnostic intent of the item.

\subsection{Clarification output}
\label{app:clarify_output}

\begin{promptbox}{Clarification Output}

\texttt{needs\_clarification}: \texttt{true}

\medskip
\textbf{Reasoning.}
\textit{``Sensor length measurement'' is vague; clarified as a displacement/position measurement sensor. Also clarified that the question asks which failure mode is NOT relevant.}

\medskip
\textbf{Clarified question.}
For a fan, if a displacement/position sensor (length measurement, e.g., axial or radial displacement) shows an abnormal reading, which of the following failure events is NOT relevant to (i.e., cannot be indicated by) this sensor?

\medskip
\textbf{Clarified options.}

\smallskip
\begin{tabularx}{\linewidth}{
    >{\bfseries}p{7mm}
    X
}
A & Damaged oil seals \\[0.8mm]
B & Bearing wear \\[0.8mm]
C & Bearing damage \\[0.8mm]
D & Fouled rotor \\[0.8mm]
E & Shaft misalignment
\end{tabularx}

\end{promptbox}

The rewrite acts on both ambiguities. The sensor is grounded as a displacement/position probe measuring axial or radial displacement, and the negative polarity is restated explicitly as ``cannot be indicated by this sensor''. The answer options are only lightly normalised: their number, order, and meaning are preserved, and no answer is suggested by the rewrite.

\newcommand{\stepnum}[1]{%
  \tikz[baseline=(n.base)]\node[circle,fill=white,text=promptheader,
    inner sep=1.2pt,minimum size=4.2mm,font=\footnotesize\bfseries](n){#1};}

\section{Full Inference Trace}
\label{app:full_trace}

This appendix reports the complete output of every stage of the online CoP pipeline for a single FailureSensorIQ item. The item belongs to the aero gas turbine subset and was processed by the full CoP configuration with Claude Opus 4.7. The prompt templates used at each step are given in Appendix~\ref{app:prompts}.

\begin{promptbox}{\stepnum{1}\; Question}

\textbf{Stem.}
For aero gas turbine, if a failure event air inlet blockage occurs, which sensor out of the choices is the most relevant sensor regarding the occurrence of the failure event?

\medskip
\begin{tabularx}{\linewidth}{
    >{\bfseries}p{7mm}
    X
    >{\raggedleft\arraybackslash}p{15mm}
}
A & vibration & \\[0.8mm]
B & oil leakage/ consumption & \\[0.8mm]
\textcolor{answerblue}{C} &
\textbf{\textcolor{answerblue}{compresor pressure/ pressure ratio}} &
\textbf{\textcolor{answerblue}{Gold}} \\[0.8mm]
D & fuel pressure/ fuel flow & \\[0.8mm]
E & power turbine temperature &
\end{tabularx}

\end{promptbox}

\begin{promptbox}{\stepnum{2}\; Clarify \hfill \normalfont\itshape Is the question ambiguous or unclear?}

\texttt{needs\_clarification}: \texttt{false} \qquad \texttt{clarified\_options}: \texttt{[]}

\medskip
The question is deemed unambiguous, so no changes are made. The original stem and options are passed unchanged to steps 3a and 3b.

\end{promptbox}

\begin{promptbox}{\stepnum{3a}\; Retrieve \& expand \hfill \normalfont\itshape What reasoning examples exist in the CKG?}

\textbf{Retrieval seeds.} The question and options are embedded and compared with the failure-mode and sensor entities of the aero gas turbine CKG. The top-2 seeds are:

\smallskip
\begin{center}
\footnotesize
\begin{tabular}{clc}
\toprule
\textbf{Rank} & \textbf{Seed entity} & \textbf{Cosine similarity} \\
\midrule
1 & Turbine inlet temperature sensor & 0.6941 \\
2 & turbine inlet temperature sensor & 0.6923 \\
\bottomrule
\end{tabular}
\end{center}

\textbf{Expanded CKG chains.} Backward expansion from the two sensor seeds yields nine complete chains. Each chain is written as failure mode $\rightarrow$ physical change(s) $\rightarrow$ \textbf{sensor}, where the arrows correspond, in order, to the CKG relations \textit{results in}, \textit{leads to}, and \textit{is monitored by}:

\smallskip
{\footnotesize\raggedright
\begin{enumerate}
\setlength\itemsep{1mm}
\item Variable stator overheating $\rightarrow$ Stator temperature at variable stators increases $\rightarrow$ Turbine efficiency at power turbine decreases $\rightarrow$ Turbine inlet temperature at turbine inlet increases $\rightarrow$ \textbf{Turbine inlet temperature sensor}
\item High turbine inlet temperature causes excess NOx $\rightarrow$ Turbine inlet temperature at turbine inlet increases $\rightarrow$ \textbf{Turbine inlet temperature sensor}
\item build-up of deposits on compressor blading $\rightarrow$ compressor isentropic efficiency at compressor decreases $\rightarrow$ turbine operating temperature at turbine increases $\rightarrow$ \textbf{turbine inlet temperature sensor}
\item excessive accessory power demand $\rightarrow$ fuel flow at combustor increases $\rightarrow$ turbine inlet temperature at turbine inlet increases $\rightarrow$ \textbf{turbine inlet temperature sensor}
\item turbine inlet temperature exceeds safe limit $\rightarrow$ turbine inlet temperature at turbine inlet increases $\rightarrow$ \textbf{turbine inlet temperature sensor}
\item nozzle area reduction $\rightarrow$ nozzle area at propelling nozzle decreases $\rightarrow$ turbine inlet temperature at turbine inlet increases $\rightarrow$ \textbf{turbine inlet temperature sensor}
\item Turbine inlet temperature exceeds limit $\rightarrow$ turbine inlet temperature at turbine inlet increases $\rightarrow$ \textbf{turbine inlet temperature sensor}
\item turbine overheating $\rightarrow$ turbine inlet temperature increases $\rightarrow$ \textbf{turbine inlet temperature sensor}
\item flame-out during rapid climb $\rightarrow$ fuel flow rate at fuel metering unit decreases $\rightarrow$ combustor temperature at combustor chamber decreases $\rightarrow$ turbine inlet temperature at turbine inlet decreases $\rightarrow$ \textbf{turbine inlet temperature sensor}
\end{enumerate}
}

\end{promptbox}

\begin{promptbox}{\stepnum{3b}\; Extract failure mode \hfill \normalfont\itshape What failure mode must be analysed?}

\textbf{Output.} \texttt{\{"failure\_modes": ["air inlet blockage"]\}}

\medskip
One failure mode is extracted from the question and options. This step does not access the CKG.

\end{promptbox}

\begin{promptbox}{\stepnum{4}\; Reason about the failure mode \hfill \normalfont\itshape What physical chains could this failure mode produce?}

\textbf{Input.} Asset (\textit{aero gas turbine}), target failure mode (\textit{air inlet blockage}), the nine CKG chains from step 3a as in-context examples, and the chain pattern. The question and options are not provided.

\medskip
\textbf{Output.} One LLM call produces ten candidate chains, none of which is copied from step 3a:
 
\smallskip
{\footnotesize\raggedright
\begin{enumerate}
\setlength\itemsep{1mm}
\item air inlet blockage $\rightarrow$ inlet air mass flow at compressor inlet decreases $\rightarrow$ compressor pressure ratio at compressor decreases $\rightarrow$ compressor discharge pressure sensor
\item air inlet blockage $\rightarrow$ inlet air pressure at compressor inlet decreases $\rightarrow$ inlet pressure sensor
\item air inlet blockage $\rightarrow$ inlet air mass flow at compressor inlet decreases $\rightarrow$ air-fuel ratio at combustor decreases $\rightarrow$ turbine inlet temperature at turbine inlet increases $\rightarrow$ turbine inlet temperature sensor
\item air inlet blockage $\rightarrow$ compressor inlet pressure at compressor inlet decreases $\rightarrow$ compressor surge margin at compressor decreases $\rightarrow$ compressor pressure sensor
\item air inlet blockage $\rightarrow$ inlet air mass flow at compressor inlet decreases $\rightarrow$ engine thrust at nozzle decreases $\rightarrow$ thrust sensor
\item air inlet blockage $\rightarrow$ pressure drop across inlet at air intake increases $\rightarrow$ differential pressure sensor
\item air inlet blockage $\rightarrow$ compressor rotational speed at compressor decreases $\rightarrow$ shaft speed sensor (N1/N2)
\item air inlet blockage $\rightarrow$ inlet air mass flow at compressor inlet decreases $\rightarrow$ exhaust gas temperature at turbine exhaust increases $\rightarrow$ exhaust gas temperature sensor
\item air inlet blockage $\rightarrow$ compressor inlet temperature at compressor inlet changes $\rightarrow$ inlet temperature sensor
\item air inlet blockage $\rightarrow$ inlet air mass flow at compressor inlet decreases $\rightarrow$ compressor stall margin at compressor decreases $\rightarrow$ vibration at compressor increases $\rightarrow$ vibration sensor
\end{enumerate}
}

\end{promptbox}

\begin{promptbox}{\stepnum{5}\; Judge \hfill \normalfont\itshape Sees question, options and all inferred chains}

\textbf{Input.} Original question and options, the task polarity instruction (\textit{MOST relevant: select the option with the strongest, most direct diagnostic relationship}), and the ten chains from step 4.

\medskip
\textbf{Judge reasoning.}
\textit{``Air inlet blockage most directly reduces inlet air mass flow and pressure, which immediately manifests as a drop in compressor pressure/pressure ratio. Multiple diagnostic chains point to compressor pressure as the primary monitored parameter. EGT/turbine temp changes and vibration are secondary effects, while oil and fuel are unrelated.''}

\end{promptbox}

\begin{promptbox}{Answer}

\textbf{\textcolor{answerblue}{C --- compresor pressure/ pressure ratio}} \hfill \textbf{\textcolor{answerblue}{Correct}}

\end{promptbox}

The trace shows the same behaviour as the fan example in Figure~\ref{fig:cop_reasoning_example}. The retrieved CKG chains concern different failure mechanisms and point to a sensor that is not the correct answer. They nevertheless provide the failure--physical change--sensor structure that the reasoning step applies to air inlet blockage. The correct answer follows from the newly inferred chains.

\end{document}